\documentclass{article}
\usepackage{iclr2027_conference,times}
\iclrfinalcopy
\usepackage[T1]{fontenc}
\usepackage[utf8]{inputenc}
\usepackage{amsmath,amssymb,amsthm,booktabs,tabularx,array,colortbl}
\usepackage{microtype,graphicx}
\usepackage{hyperref}
\hypersetup{hypertexnames=false,hidelinks,
  pdftitle={EPOCH: Reliable Discovery through Evidence-Governed Search},
  pdfauthor={Binjie Guo, Aisheng Mo, Ruitong Li, Xinle Deng}}
\usepackage{xurl}
\usepackage{placeins}
\usepackage{float}
\newcolumntype{Y}{>{\raggedright\arraybackslash}X}
\newcolumntype{R}{>{\raggedleft\arraybackslash}X}
\newcolumntype{L}[1]{>{\raggedright\arraybackslash}p{#1}}
\newcolumntype{B}[1]{>{\bfseries\raggedright\arraybackslash}p{#1}}
\newcommand{\epoch}{\textsc{EPOCH}}
\newcommand{\code}[1]{\nolinkurl{#1}}
\newcommand{\EPOCHSharedPreamble}{1}
\DeclareMathOperator{\Cov}{Cov}
\DeclareMathOperator{\tr}{tr}
\DeclareMathOperator{\brank}{\underline{R}}
\definecolor{EpochBlue}{HTML}{235BB5}
\definecolor{EpochBluePale}{HTML}{EAF1FC}
\definecolor{EpochTeal}{HTML}{087F78}
\definecolor{EpochTealPale}{HTML}{E9F6F3}
\definecolor{EpochSlatePale}{HTML}{F3F6FA}

\title{EPOCH: Reliable Discovery through Evidence-Governed Search}
\author{Binjie Guo$^{1}$ \quad Aisheng Mo$^{1}$ \quad Ruitong Li$^{2}$ \quad Xinle Deng$^{1}$\\[0.5em]
\normalfont $^{1}$Zhejiang University\\
\normalfont $^{2}$The University of Hong Kong}

\begin{document}
\maketitle
\lhead{EPOCH: arXiv preprint}
\raggedbottom

\begin{abstract}
AI research agents are increasingly used to search over programs, mathematical constructions, and proofs. However, existing systems typically optimize evaluator feedback without adequately governing how that feedback is interpreted, challenged, and reused. As a result, promising but fragile candidates can be promoted as discoveries, while benchmark improvements, finite certificates, and theorem-level claims are too easily conflated. We introduce EPOCH, an evidence-governed architecture designed to close this gap. EPOCH implements an evidence-governed discovery loop by combining explicit task contracts, typed memory, active falsification, admission checks, and independent replay, so that each candidate is evaluated against the strength and scope of the claim it supports. EPOCH achieves state-of-the-art aggregate performance on AlgoTune, substantially exceeding the strongest baseline in mean normalized score (0.65 vs. 0.53), and attains the highest mean score on the internal Math14 suite (0.57). It further shows favorable held-out behavior under official-test replay and leads the descriptive aggregate on AgentHPO. Across ten discovery problems, EPOCH delivers substantial task-specific advances, including improved executable constructions, optimized algorithms, counterexamples, and proof-supported results. These advances demonstrate its ability to convert search into concrete progress across mathematical and computational domains. Together, the results suggest that evidence governance is a necessary step toward AI research agents that produce not only stronger solutions, but also more trustworthy scientific discoveries.
\end{abstract}

\section{Introduction}

AI research agents are increasingly being developed not only to solve well-specified tasks, but also to generate candidate discoveries in code, algorithms, mathematics, and science. Recent systems have demonstrated this potential across games, program synthesis, tensor decompositions, geometric constructions, formal proofs, and scientific screening \citep{alphazero,muzero,alphacode,alphatensor,alphageometry,alphaproof,gnome}. FunSearch and AlphaEvolve further show that evaluator-guided evolution can transform large language models into effective engines for open-ended search \citep{funsearch,alphaevolve}. Despite this progress, current systems are primarily designed to generate and optimize candidates, rather than to construct the evidence needed to support a discovery. A promising program, construction, or conjecture is only a starting point: establishing its value may require identifying a mechanism, producing a counterexample to a competing explanation, checking a finite certificate, or distinguishing an experimentally supported observation from a proof-backed claim. These supporting objects are often generated, tested, and recorded only opportunistically. As a result, an agent may return an improved candidate without a clear account of why it works, what evidence supports it, or how that evidence can be independently revisited. The central challenge is therefore to couple search with an explicit process for constructing, challenging, and retaining the evidence that turns promising candidates into supported discoveries.

Existing agents combine generation with execution, testing, or verification, but they do not provide a common protocol for this evidence-construction process across domains. We ask: \emph{what protocol should govern the promotion of a candidate into a retained discovery when the required supporting evidence differs across tasks?} For code, this may involve semantic checks, resource measurements, and held-out replay; for algebra, exact identities, constructions, or obstruction witnesses; and for theorem discovery, computational evidence and formal proof with distinct scopes. A scalar reward can guide a search procedure toward better candidates, but it cannot by itself organize the generation of supporting evidence, specify the scope of a claim, or preserve the failed attempts that may explain or challenge it.

\begin{figure}[t]
\centering
\includegraphics[width=0.84\linewidth]{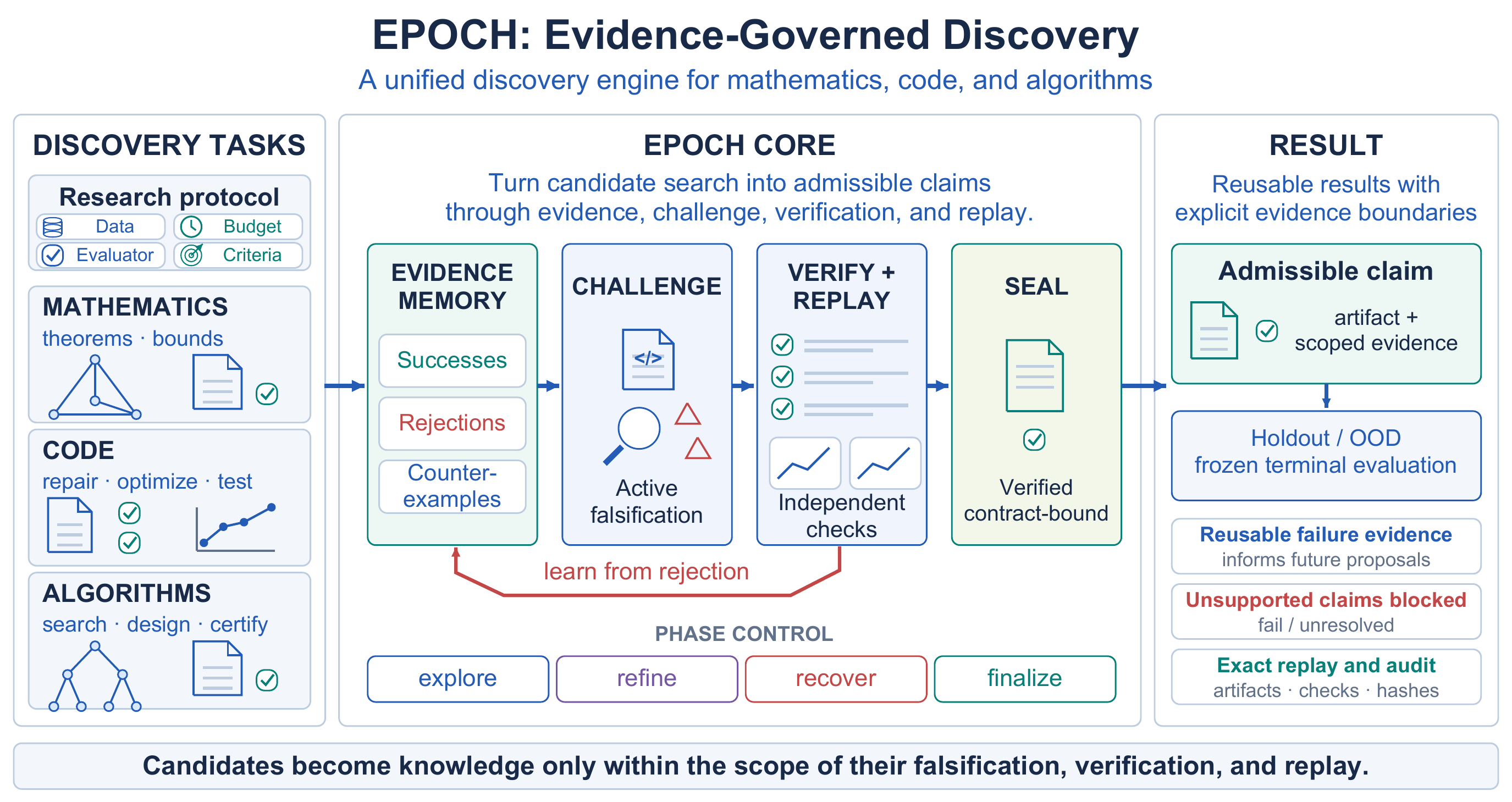}
\caption{\epoch{} couples discovery search to an explicit evidence ledger. Successful and failed trials enter typed memory; failures are fed back into search for repair and refinement (red loop). Challenge, admission, and independent replay determine whether a candidate is retained, while the task contract fixes the checks, budget, and claim scope governing that decision.}
\label{fig:concept}
\end{figure}

We propose \epoch{} (Evidence-conditioned Proposal Optimization through Cumulative Hypothesis-governance, Fig.~\ref{fig:concept}), an evidence-governed architecture that frames discovery as constrained search over an explicit evidence ledger. Each task begins with a contract specifying permissible edits, applicable checks, resource limits, and the scope of admissible claims. Typed memory retains both successful and failed trials, allowing later search to reuse supporting evidence and avoid known failure modes. Candidates are actively challenged against their declared mechanisms, admitted only when their support satisfies the task contract, and independently replayed after the search endpoint is frozen. Rather than making a verifier complete or elevating finite evidence into a general theorem, \epoch{} provides a domain-general protocol that aligns each retained result with its supporting evidence and makes the limits of that support explicit.

Our contributions are:
\vspace{\baselineskip}
\begingroup
\setlength{\topsep}{0pt}
\setlength{\partopsep}{0pt}
\setlength{\itemsep}{0pt}
\setlength{\parsep}{0pt}
\setlength{\parskip}{0pt}
\begin{itemize}
\item \textbf{Evidence-governed discovery.} We introduce a domain-general architecture in which task contracts, typed memory, active falsification, claim-aware admission, and independent replay jointly govern how candidate solutions become retained results.
\item \textbf{Strong exploration and exploitation performance.} \epoch{} achieves state-of-the-art aggregate performance on AlgoTune, the highest mean score on internal Math14, favourable official-test replay behaviour, and the leading descriptive aggregate on AgentHPO.
\item \textbf{Major discoveries with scoped evidence.} Across ten discovery problems, \epoch{} delivers eleven improved tensor border-rank upper bounds, including an explicit 30-path construction for $T_{\rm ss444}$; a reusable uniform Big-centroid construction; a record-setting 54-comparison executable procedure for median-of-23 selection; a 1,512-gate ($14.10\%$) matched-PyZX T-count reduction for AES; a Lean-checked all-dimensional hypercube determinant inequality with complete equality classification; and a Lean-checked nine-element counterexample to the proposed $H_\beta$ coefficient ordering, extended to an infinite family.
\end{itemize}
\endgroup

\section{Related Work}

\paragraph{Evaluator-Guided Search and Discovery.}
AlphaCode generates programs at scale and filters, clusters, and reranks them according to their behavior on generated test inputs, whereas AlphaTensor formulates matrix-multiplication algorithm discovery as a search for low-rank decompositions of the matrix-multiplication tensor \citep{alphacode,alphatensor,alphatensorcode}. FunSearch couples language-model proposals to evolutionary selection, and AlphaEvolve extends evaluator-guided evolution to mathematical and computational tasks \citep{funsearch,alphaevolve}. OpenEvolve, AdaEvolve, and AI-Scientist further broaden the landscape of evolutionary and end-to-end research workflows \citep{openevolve,adaevolve,aiscientist1,aiscientist2}. These systems establish the effectiveness of evaluator-guided search; \epoch{} complements them by introducing an evidence contract that governs which candidates may be retained as discoveries, what support their claims require, and how rejected paths remain available for subsequent search.

\paragraph{Domain-Specific Evidence for Discovery.}
AlphaGeometry combines learned construction with symbolic deduction, AlphaProof searches formally checkable Lean states, and GNoME couples learned screening to first-principles calculations \citep{alphageometry,alphageometrycode,alphaproof,gnome,gnomecode}. Their domain-specific validators are essential rather than interchangeable. Recent AI-assisted mathematical studies likewise combine model-guided exploration with expert scrutiny and, where applicable, formal verification, while distinguishing suggestive evidence from established mathematical claims \citep{clauderiemann,navierstokes2026}. \epoch{} therefore does not seek to replace specialist validation with a universal truth score; instead, it standardizes how domain-specific outputs support discovery. Typed receipts bind a retained result to the checks that produced its evidence, three-valued gates distinguish supported, refuted, and unresolved candidates, and independent replay tests frozen artifacts outside the search loop. The result is a replayable admission record that specifies what was checked, what outcome was obtained, and which claim the available evidence permits.

\section{Evidence-Governed Discovery}

We organize the protocol around three decisions: what counts as an admissible endpoint, how evidence changes subsequent proposals, and what may be claimed after independent replay.

\subsection{Task contract and admissible endpoints}

A task contract is $\mathcal C=(D_{\rm pub},D_{\rm term},E,B,A,S)$: public and terminal data views, evaluators, resource budget, editable artifact space, and allowed claim scope. A candidate $c$ has objective $q_{\mathcal C}(c)$ and evidence bundle $\xi$; a claim $\sigma$ states what the result asserts. Endpoint selection is
\begin{equation}
\label{eq:constrained}
\begin{gathered}
(c^*,\xi^*,\sigma^*)\in\arg\max_{c,\xi,\sigma}\ q_{\mathcal C}(c)\\
\begin{aligned}
\text{s.t.}\quad &c\in A,\quad u_j(c,\xi)\le B_j,\\
&\operatorname{Correct}_E(c;\xi)=1,\quad
\operatorname{Challenge}(c;\xi)=1,\\
&\operatorname{Replay}_E(c;\xi)=1,\quad
\operatorname{Supports}(\xi,\sigma)=1,\quad \operatorname{Scope}(\sigma)\subseteq S.
\end{aligned}
\end{gathered}
\end{equation}
The maximization ranges only over proposals generated within the registered run, including a verified fallback when supplied. It is not a global optimum over all programs or proofs. Gates take values in $\{1,0,\bot\}$, where $\bot$ denotes unresolved evidence; only 1 permits admission. 

\subsection{Evidence memory, falsification, and phase control}

At step $t$, the state is $s_t=(c_t^*,M_t,z_t,b_t)$: incumbent, append-only memory, phase, and remaining resources. The proposal policy emits
\begin{equation}
a_t=(c_t,H_t,\widehat\delta_t,R_t,F_t)
\sim\pi_\theta(\cdot\mid\mathcal C,\mathcal V(M_t),z_t,b_t),
\end{equation}
containing a candidate, hypothesis, predicted effect, evidence references, and intended falsifier. The bounded view $\mathcal V(M_t)$ exposes relevant prior evidence without rewriting the ledger. Each observation records artifact and evaluator hashes, score, correctness, resource use, failure attribution, and rejection reason. Scientific failure can therefore constrain a later proposal, whereas an infrastructure incident can trigger bounded recovery without masquerading as evidence against the method.

A challenge targets the stated mechanism rather than merely repeating the objective. Examples include semantic boundary inputs, an algebraic coefficient mismatch, a cokernel witness, counterexample search, or a proof obligation. Passing a finite challenge supports only its declared coverage. The controller chooses among explore, refine, recover, and finalize; it does not imply an unreported learned optimal schedule. If $\mathcal A_t$ contains candidates whose applicable gates pass, the retained endpoint satisfies $c_t^*\in\arg\max_{c\in\mathcal A_t}q_{\mathcal C}(c)$. With a verified fallback, this gives a protocol-level safety invariant, not correctness beyond evaluator coverage.

\subsection{Replay and claim sealing}

Terminal replay evaluates a frozen selected endpoint and cannot feed back into selection. A sealed result binds the contract, endpoint, ledger, evidence, claim, and scope:
\begin{equation}
\operatorname{Seal}=h(\mathcal C,c^*,M_T,\xi^*,\sigma^*,S).
\end{equation}
Independent implementations may reconstruct semantics rather than rerun the same checker. For mathematical outputs, the bundle distinguishes finite computation, general argument, formal kernel, and missing bridge. 

\section{Evaluation}

We first compare endpoint quality and held-out behavior, then examine component removals and the distribution of performance across tasks. The discovery audit in Section~\ref{sec:portfolio} asks a separate question: what reusable objects and bounded claims emerge from this process?

\subsection{Benchmarks and Estimands}

We evaluate \epoch{} in three complementary settings: AlgoTune for constrained algorithmic search, Math14 for diverse mathematical objectives with exact endpoint checkers, and six AgentHPOBench tasks for component ablations \citep{algotune,agenthpo}. On a frozen 12-task AlgoTune subset, five methods receive the same budget of 48 semantic calls and are run with three seeds; we report mean task scores and their macro-average. To test whether search-time gains transfer beyond development data, we freeze each eligible final program and replay it, without further search or adaptation, on 32 previously unused official test inputs. This replay reruns correctness checks and fresh-process timing. Math14 evaluates performance across 14 mathematical tasks, whereas AgentHPO isolates the contribution of EPOCH's individual components. Full task-level results, seeds, measurements, and statistical procedures are provided in the Supplement.

\subsection{Search Performance and Held-Out Robustness}

\epoch{} achieves the highest AlgoTune development mean (Table~\ref{tab:benchmarks}), reaching 0.65 and surpassing the strongest baseline, AdaEvolve, by $22.60\%$. It also exceeds both AI-Scientist variants and OpenEvolve by margins ranging from $24.80\%$ to $30.90\%$, establishing the strongest aggregate performance among the evaluated methods. This advantage extends to the held-out official-test replay: after endpoints are frozen and evaluated without further search or adaptation, EPOCH has positive mean effects against all four baselines. Maximum common subgraph retains a $+0.48$ normalized-score advantage and Convolve2D a $+0.24$ advantage, although NMF reverses to $-0.27$. The largest development gains occur on Convolve2D, maximum common subgraph, Robertson ODE, NMF, and zeta-zero count, reflecting task-specific improvements to the evaluated executable endpoints under the benchmark protocol.

\begin{table}[t]
\centering
\small
\caption{AlgoTune scores. Development uses 12 tasks and three seeds per method; replay effects use seven official-test tasks. Higher scores are better. Relative gain is measured against the named method's development score. Replay effects are \epoch{} minus the named baseline in normalized-score units; the development and replay columns use distinct endpoint populations.}
\label{tab:benchmarks}
\begin{tabularx}{\linewidth}{@{}lRRR@{}}
\toprule
\rowcolor{EpochBluePale}
Method & Dev. score & Rel. gain & Replay effect \\
\midrule
\rowcolor{EpochTealPale}
\textbf{\epoch{}} & \textbf{0.65} & \textbf{Reference} & \textbf{Reference} \\
AdaEvolve & 0.53 & $\textcolor{EpochBlue}{+22.60\%}$ & $\textcolor{EpochBlue}{+0.04}$ \\
\rowcolor{EpochSlatePale}
AI-Scientist v1 & 0.52 & $\textcolor{EpochBlue}{+24.80\%}$ & $\textcolor{EpochBlue}{+0.06}$ \\
OpenEvolve & 0.50 & $\textcolor{EpochBlue}{+30.10\%}$ & $\textcolor{EpochBlue}{+0.09}$ \\
\rowcolor{EpochSlatePale}
AI-Scientist v2 & 0.50 & $\textcolor{EpochBlue}{+30.90\%}$ & $\textcolor{EpochBlue}{+0.11}$ \\
\bottomrule
\end{tabularx}
\end{table}

On Math14 (Table~\ref{tab:math14}), \epoch{} achieves the highest mean score of 0.57, outperforming every evaluated baseline. Its relative improvement ranges from $0.60\%$ over OpenEvolve to $10.90\%$ over AI-Scientist v1, demonstrating that EPOCH's evidence-governed search architecture transfers effectively from executable optimization to diverse mathematical objectives.

\begin{table}[t]
\centering
\small
\caption{Math14 internal transfer scores over 14 objectives. Higher scores are better; relative gains are measured against the named baseline.}
\label{tab:math14}
\begin{tabularx}{\linewidth}{@{}lRRR@{}}
\toprule
\rowcolor{EpochBluePale}
Method & Mean score & Rel. gain & Rank \\
\midrule
\rowcolor{EpochTealPale}
\textbf{\epoch{}} & \textbf{0.57} & \textbf{Reference} & \textbf{1} \\
OpenEvolve & 0.56 & $\textcolor{EpochBlue}{+0.60\%}$ & 2 \\
\rowcolor{EpochSlatePale}
AdaEvolve & 0.53 & $\textcolor{EpochBlue}{+7.70\%}$ & 3 \\
AI-Scientist v2 & 0.52 & $\textcolor{EpochBlue}{+9.20\%}$ & 4 \\
\rowcolor{EpochSlatePale}
AI-Scientist v1 & 0.51 & $\textcolor{EpochBlue}{+10.90\%}$ & 5 \\
\bottomrule
\end{tabularx}
\end{table}

\subsection{Component contributions and task-level trade-offs}

Whereas the main benchmarks compare complete systems, six AgentHPOBench tasks allow us to isolate the contributions of \epoch{}'s major components. Because these tasks use heterogeneous metrics, task-level raw scores remain the primary results. For descriptive cross-task comparison, we normalize each task-level mean $z_{tv}$ within its task, accounting for whether the task metric is maximized or minimized:

\begin{equation}
N_{tv}
=
100
\frac{
z_{tv}-\min_{v'} z_{tv'}
}{
\max_{v'} z_{tv'}-\min_{v'} z_{tv'}
},
\qquad
\bar N_v
=
\frac{1}{6}\sum_t N_{tv}.
\end{equation}
Thus, $\bar N_v$ summarizes the relative performance of each variant across the six tasks without conflating their native metrics.

\begin{table}[!b]
\centering
\small
\caption{Ablations on six AgentHPOBench tasks. Full \epoch{} achieves the highest descriptive normalized mean. Removing any component reduces this mean by 13.17--55.59 points. W/T/L denotes the task-level record of Full \epoch{} against each ablation in the paired report.}
\label{tab:ablation}
\begin{tabular}{@{}lrrc@{}}
\toprule
Variant & Norm. mean & Gap & Full W/T/L \\
\midrule
Full \epoch{}       & \textbf{77.48} & --        & --    \\
No memory           & 64.32          & $-13.17$ & 4/0/2 \\
No proof admission  & 40.21          & $-37.28$ & 6/0/0 \\
No falsifier        & 35.74          & $-41.75$ & 4/1/1 \\
No phase controller & 21.89          & $-55.59$ & 5/0/1 \\
\bottomrule
\end{tabular}
\end{table}

Full \epoch{} achieves both the highest normalized mean (Table~\ref{tab:ablation}) and the best mean task rank (1.67). Removing memory lowers the normalized mean by 13.17 points, while removing proof admission, falsification, or phase control produces larger reductions of 37.28, 41.75, and 55.59 points, respectively. The paired comparisons likewise favor the complete system over every ablation, including a 6/0/0 record against the variant without proof admission. The no-memory variant performs better on ART and ModernBERT, consistent with an exploration--exploitation trade-off: unconstrained proposals can benefit short-horizon tuning on individual tasks, whereas memory improves performance across the broader suite by preserving evidence and steering search away from previously unsuccessful regions. This pattern highlights the complementary roles of EPOCH's components rather than a uniform per-task effect. The broader findings are supported independently: AlgoTune and Math14 evaluate the complete architecture end to end, while the case-study results are established through task-specific certificates and replay. Overall, the complete architecture achieves the strongest performance across the broadest range of tasks.

\begin{figure}[H]
\centering
\includegraphics[width=\linewidth]{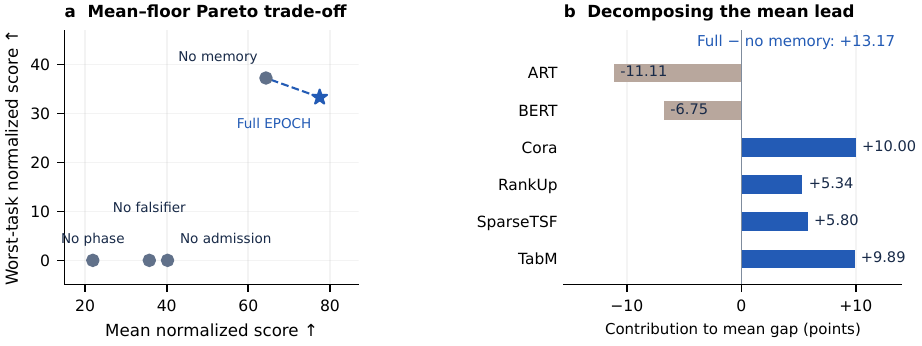}
\caption{Performance trade-offs across AgentHPOBench ablations. \textbf{a}, Full \epoch{} achieves the highest normalized mean, whereas the no-memory variant attains the highest minimum task score; the dashed segment connects the observed nondominated variants. \textbf{b}, Task-level decomposition of the 13.17-point normalized-mean gap between Full \epoch{} and the no-memory variant. Contributions sum to the aggregate gap.}
\label{fig:mechanism-traces}
\end{figure}

\paragraph{From average performance to task coverage.}
Performance profiles test whether aggregate gains extend across tasks rather than arising from a small number of outliers. On AlgoTune, \epoch{} scores at least $0.60$ on five of twelve tasks, compared with at most one task for any baseline. On AgentHPOBench, it reaches a normalized score of at least $80$ on three of six tasks, compared with at most two for any ablation. The advantage of the complete system therefore extends beyond its mean score to broader coverage of high-performing tasks.

\begin{figure}[H]
\centering
\includegraphics[width=\linewidth]{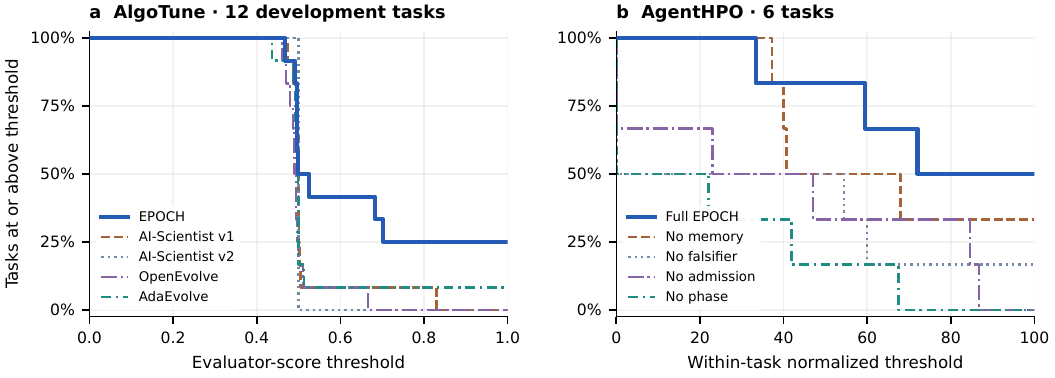}
\caption{Empirical performance profiles, $P_v(\tau)=|\{t:s_{tv}\geq\tau\}|/T$, showing the fraction of tasks on which variant $v$ reaches threshold $\tau$. Higher curves indicate broader coverage at a given performance level. Left, twelve AlgoTune development tasks. Right, six AgentHPOBench tasks, using task-level means normalized as in Eq.~(4). The panels use different score scales.}
\label{fig:performance-profiles}
\end{figure}
\FloatBarrier
\section{From Search to Discovery: Case Studies}
\label{sec:portfolio}

\begin{figure}[!htbp]
\centering
\includegraphics[width=\linewidth]{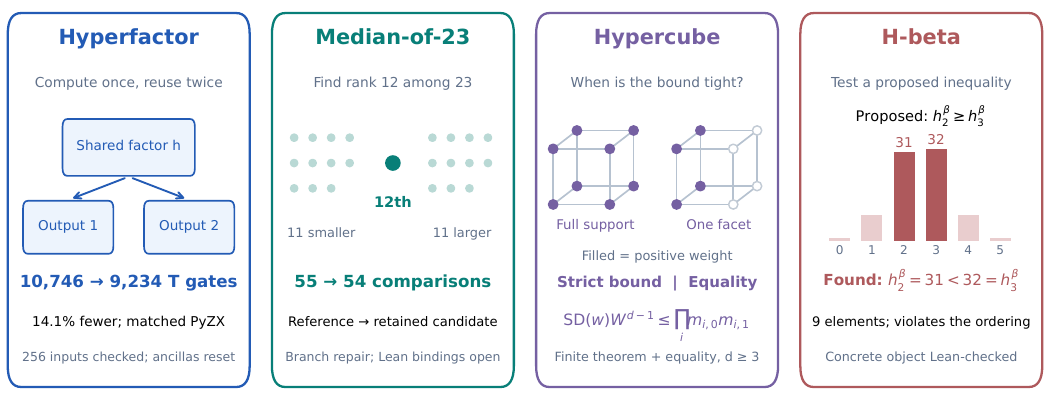}
\caption{Four discovery cases, left to right. Hyperfactor reuses one factor across outputs, reducing circuit cost. The median has eleven smaller and eleven larger elements; branch repair yields the comparison bound. Full cube support gives a strict inequality; facet support gives equality (filled vertices: positive weight). H-beta violates the proposed coefficient ordering. Schematics are not emitted circuits or decision trees; Table~\ref{tab:portfolio} records evidence boundaries.}
\label{fig:case-atlas}
\end{figure}

The benchmark results establish EPOCH's search performance; we next examine the discoveries that this search produces. The following ten case studies span executable optimization, algorithmic construction, formal statements, and counterexamples, illustrating how EPOCH couples candidate generation with task-specific evidence. Figure~\ref{fig:case-atlas} highlights four representative outcomes, while Table~\ref{tab:portfolio} summarizes the retained results, their supporting evidence, and the scope of each claim. We begin with executable constructions and repair, and then turn to formal results and counterexamples; complete derivations are provided in the supplementary.

\FloatBarrier

\begin{table}[t]
\centering
\footnotesize
\caption{Documented discovery outcomes and their evidential scope.}
\label{tab:portfolio}
\begin{tabularx}{\linewidth}{@{}B{0.18\linewidth}YL{0.31\linewidth}@{}}
\toprule
\rowcolor{EpochBluePale}
Outcome & Retained result & Evidence boundary \\
\midrule
\rowcolor{EpochTealPale}
2-adic transfer & 11 border-bound improvements, including $\brank(T_{\rm ss444})\le30$ & 45,242 exact coordinates; Lean checks flagship coefficients\\
Hyperfactor & AES sharing: 10,746 to 9,234 T gates under matched PyZX & All 256 inputs and ancillas checked \\
\rowcolor{EpochSlatePale}
Big-centroid & Uniform $q+u_1$-term constructor under stated field hypotheses & 278,904 identities; general existence/lower-bound bridge separated \\
Median-of-23 & Executable candidate $V_{12}(23)\le54$ & Complete census and repair policies \\
\rowcolor{EpochSlatePale}
Hypercube & All-dimensional determinant inequality with equality classification & Full finite-polynomial theorem Lean-checked; geometric alignment separate \\
H-beta & Nine-element counterexample and stable extension-family defect $-1$ & Concrete object Lean-checked \\
\rowcolor{EpochSlatePale}
Three-state kernel & Exact nonnegativity candidate for the stated reversible kernel & Lean mixing closure \\
Full Perazzo & Candidate $\brank=q+n$ and $R_{\rm W}=dq$ & Kernel lemmas formalized \\
\rowcolor{EpochSlatePale}
Heisenberg & Principal-chain limit $m^{1/(p+1)}$ versus determinant $m$ & Finite implication formalized \\
Online embedding & Alternating-family leading constant and $k\log k$ correction & Critical identities formalized \\
\bottomrule
\end{tabularx}
\end{table}

\subsection{Constructive Discoveries and Algorithmic Optimization}

\paragraph{Characteristic-two tensor transfer.}
Low-border-rank decompositions are fundamental building blocks for fast bilinear algorithms: each explicit construction provides a concrete certificate of reduced algebraic complexity. A central difficulty is that rational decompositions can collapse after reduction to characteristic two, preventing their direct use over the target field. \epoch{} turns this obstruction into a constructive search signal: it extracts the incompatible coefficient constraints, uses their witnesses to select a new degeneration chart, and searches for a valid characteristic-two replacement. Across the audit, this procedure transfers all 43 directly applicable formulas and covers 24 coordinate-distinct targets, yielding eleven improved border bounds. In particular, it discovers a 30-path construction for $T_{\rm ss444}$, establishing $\brank(T_{\rm ss444})\le 30$. The result contributes explicit low-complexity decompositions in a regime where naïve transfer fails, expanding the constructive toolkit available for algebraic-complexity and fast-algorithm design.

\paragraph{Uniform Big-centroid construction.}
Big-centroid constructions are particularly valuable when they extend across a family of dimensions: unlike an isolated decomposition, a uniform family can serve as a reusable object for asymptotic border-rank and algebraic-complexity analysis. \epoch{} discovers such a family, deriving an explicit $q+u_1$-term construction, where $q=\prod_{s=2}^d u_s$, from a reusable left-inverse identity rather than repeated ambient linear solves. The resulting formula applies throughout the stated parameter and field regimes, replacing a sequence of instance-specific searches with a single dimension-general construction. This shift from individual examples to an explicit scalable family provides a concrete foundation for studying the associated complexity bounds and for developing subsequent centroid-based tensor constructions.

\paragraph{Improved median-of-23 selection.}
Median selection is a fundamental problem in comparison complexity, and the classical Hadian--Sobel construction gives a 55-comparison upper bound for selecting the median of 23 elements \citep{hadian_sobel}. \epoch{} improves on this benchmark by treating over-budget execution paths as reusable counterexamples: it organizes 1,454 difficult cases into sixteen repair policies that eliminate the remaining excess comparisons. The resulting executable decision procedure has worst-case cost at most 54 comparisons, establishing $V_{12}(23)\le54$ and improving the classical bound by one comparison. This result demonstrates that failure-guided search can produce a concrete improvement to a long-standing constructive benchmark in exact selection.

\begin{figure}[!htbp]
\centering
\includegraphics[width=0.60\linewidth]{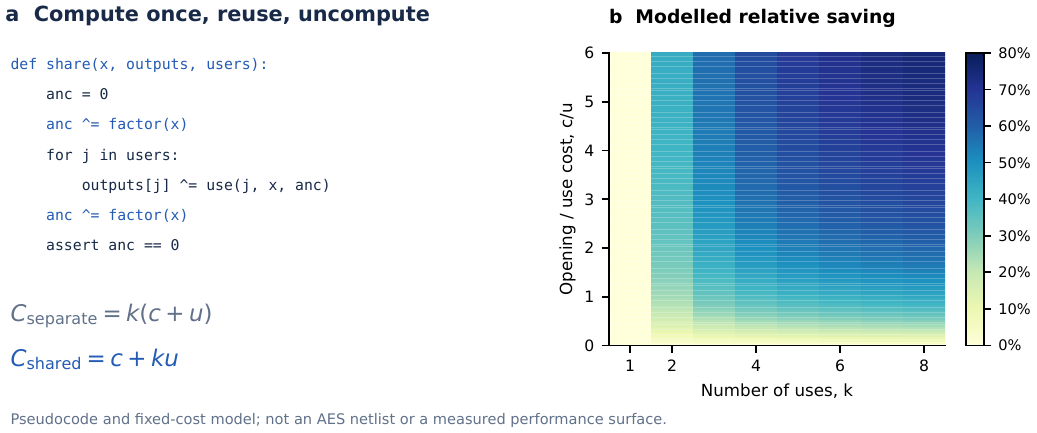}
\caption{Factor sharing and its cost scaling. A shared factor is computed once, reused across $k$ outputs, and then uncomputed. With opening cost $c$ and per-use cost $u$, the relative saving is $(k-1)c/[k(c+u)]$ and increases with the number of uses. The pseudocode preserves the input and restores the ancilla.}
\label{fig:sharing-detail}
\end{figure}

\paragraph{Reversible circuit factor sharing.}
T gates are a dominant resource in fault-tolerant quantum computation, making T-count reduction a central objective in reversible-circuit optimization \citep{pyzx}. In the evaluated AES construction, common intermediate factors are computed independently for multiple outputs, introducing substantial redundant non-Clifford cost. \epoch{} identifies these factors and restructures the circuit so that each is computed once, shared across the relevant outputs, and subsequently uncomputed. Figure~\ref{fig:sharing-detail} illustrates this mechanism and shows how the relative saving increases with the number of factor uses. Under matched PyZX synthesis, the resulting circuit reduces the T-count from 10,746 to 9,234, eliminating 1,512 T gates and lowering the non-Clifford cost by $14.10\%$, while preserving all 256 input-output cases and restoring every ancilla. This result provides a substantial resource reduction for a cryptographically relevant reversible circuit under the stated library and synthesis model.

\subsection{Mathematical Discoveries}

The remaining cases distinguish a checked general statement, an explicit counterexample, and theorem candidates with open proof bridges.

\paragraph{A sharp hypercube inequality with equality characterization.}
Let $w_v\ge0$ assign weights to the vertices $v\in\{0,1\}^d$, let $W=\sum_v w_v$, and let $m_{i,b}=\sum_{v_i=b}w_v$ denote the mass on the $b$-facet in coordinate $i$. For the unordered weighted sum $\operatorname{SD}(w)$ of absolute affine simplex determinants, EPOCH establishes
\begin{equation}
\operatorname{SD}(w)W^{d-1}\le\prod_{i=1}^d m_{i,0}m_{i,1}.
\end{equation}

For $d\ge3$, equality holds exactly when all mass is supported on a facet. Thus coordinate marginals sharply bound the global determinant quantity and characterize its degeneracies. Relating this finite-polynomial theorem to the source geometric statement still requires the stated normalization bridge \citep{hypercubesource}.

\paragraph{A stable counterexample to $H_\beta$ front-loading.}
The proposed rank-five $H_\beta$ front-loading property requires $h_2^\beta\ge h_3^\beta$. EPOCH discovers a nine-element matroid with
\begin{equation}
H_0=(1,9,27,27,9,1),\qquad H_\beta=(1,9,31,32,9,1),
\end{equation}
and hence $h_2^\beta-h_3^\beta=-1$, directly refuting the ordering. A two-line principal-extension mechanism preserves the defect for every nonnegative extension count, producing an infinite family while leaving neighboring conjectures unresolved \citep{hbetasource}.

\paragraph{Additional Mathematical Results and Verification Boundaries.}
Four further candidates concern a three-state kernel, Full Perazzo ranks, a Heisenberg limit \citep{heisenbergsource}, and online-embedding asymptotics \citep{onlinesource}. Table~\ref{tab:portfolio} summarizes their verified cores and the remaining bridges: boundary certificates, interpolation and rank arguments, representation limits, and recurrence asymptotics. Full details appear in the supplement.

\section{Discussion and Limitations}

The benchmarks and portfolio support complementary conclusions. Aggregate gains show that contracts, typed memory, falsification, admission, and replay can guide search, while the cases show how failures become repair strategies and verification turns candidates into scoped research artifacts. This evidence discipline is particularly relevant to AI4Math and AI4Science workflows combining code, symbolic reasoning, simulation, and formal methods.

Two limitations define the next frontier. First, larger paired evaluations with fixed budgets and comprehensive resource accounting are needed. Second, although EPOCH makes evidential gaps explicit, it does not resolve them all; specialist claims still require external review, statement alignment, and priority assessment. Extending the protocol to simulations, experiments, databases, scientific instruments, and expert interaction remains an important direction.

\section{Conclusion}

\epoch{} reframes AI research agents from candidate generators into systems for evidence-guided discovery. By combining typed memory, active falsification, admission, and independent replay, EPOCH turns successful, failed, and repaired search trajectories into retained results with explicit supporting evidence. Across ten discovery problems, this process yields concrete advances: eleven improved tensor border bounds, a 1,512-T-gate reduction in an AES reversible circuit, a 54-comparison median-of-23 procedure that improves the classical 55-comparison Hadian--Sobel bound \citep{hadian_sobel}, a sharp all-dimensional hypercube inequality with an equality characterization, and an infinite family refuting the proposed $H_\beta$ front-loading property. Together with state-of-the-art aggregate results on AlgoTune and leading performance on Math14 and AgentHPO, these outcomes show that evidence-governed search can do more than optimize benchmark scores: it can generate substantive algorithmic and mathematical discoveries.

\par\label{main-text-end}

\clearpage
\section*{AI Use Statement}
Generative AI tools were used to assist with manuscript polishing and with the development of candidate programs, constructions, mathematical arguments, and proofs. The authors reviewed all AI-assisted outputs. Executable results were checked through task-specific evaluators and independent replay; finite mathematical claims were verified by exact or exhaustive computation; and formal results were checked in Lean where indicated. The authors take full responsibility for the final text, claims, code, and mathematical artifacts.

\section*{Reproducibility Statement}
The supplementary material provides the information needed to audit and reproduce our reported results. For the empirical evaluations, it records frozen implementation status, task cohorts and data splits, model and seed identifiers, search budgets, correctness and timing procedures, endpoint-selection and held-out replay protocols, and task-level results. For the discovery cases, it includes detailed derivations, exact or exhaustive verification procedures, formalization scopes, outstanding proof obligations, and documented replay interfaces. All claims are restricted to the scope established by the corresponding evidence; in particular, finite certificates and formally verified kernels are not presented as establishing broader results.

{\footnotesize
\bibliography{references}
\bibliographystyle{iclr2027_conference}
}

\clearpage
\ifdefined\EPOCHSharedPreamble
\else
\documentclass{article}
\usepackage{iclr2027_conference,times}
\usepackage[T1]{fontenc}
\usepackage[utf8]{inputenc}
\usepackage{amsmath,amssymb,amsthm,booktabs,tabularx,array}
\usepackage{microtype,hyperref,xurl,graphicx}
\hypersetup{hidelinks}
\newcolumntype{Y}{>{\raggedright\arraybackslash}X}
\newcommand{\epoch}{\textsc{EPOCH}}
\newcommand{\code}[1]{\nolinkurl{#1}}
\DeclareMathOperator{\Cov}{Cov}
\DeclareMathOperator{\tr}{tr}
\DeclareMathOperator{\brank}{\underline{R}}
\def\EPOCHStandaloneSupplement{1}
\title{Supplementary Material for EPOCH:\\Evidence-Governed Discovery Across Code, Algorithms, and Mathematics}
\author{Anonymous authors}
\begin{document}
\maketitle
\ificlrfinal
  \lhead{Published as a conference paper at ICLR 2027}
\else
  \lhead{Under review as a conference paper at ICLR 2027}
\fi
\setcounter{tocdepth}{1}
\tableofcontents
\clearpage
\fi
\appendix
\setcounter{figure}{0}
\renewcommand{\thefigure}{S\arabic{figure}}
\renewcommand{\theHfigure}{supp.\arabic{figure}}
\setcounter{table}{0}
\renewcommand{\thetable}{S\arabic{table}}
\renewcommand{\theHtable}{supp.\arabic{table}}
\setcounter{equation}{0}
\renewcommand{\theequation}{S\arabic{equation}}
\renewcommand{\theHequation}{supp.\arabic{equation}}
\section{Evidence Inventory and Reporting Conventions}
\label{app:protocol}
This supplement distinguishes among original runs, the official replay, mechanism diagnostics, and documented discovery artifacts. Across the registered development matrix, EPOCH achieves the highest mean score (0.65), compared with 0.53 for the strongest baseline. In the official replay, its mean effect is positive against all four baselines, with a maximum task-level gain of approximately 0.48. On AgentHPO, the full system attains a normalized mean score of 77.48—13.17 points above the strongest component-removal variant—and proof admission is favorable on all six reported tasks. The discovery portfolio further contributes exact tensor constructions, circuit reductions, finite counterexamples, and formal kernels. The remainder of this supplement explains how these advantages arise and how each mathematical conclusion is established.

\paragraph{How to read the figures.}
Benchmark plots reproduce the supplied aggregate records. Circuit and proof diagrams explain dependencies; they are not additional measured results. Curves in Figures~\ref{fig:s-perazzo}--\ref{fig:s-online} are evaluations of the explicitly stated formulas, with hypotheses and proof status preserved. They should not be counted as independent confirmation of those formulas. Exact numerical results remain in the adjacent tables.

\subsection{Pre-registered Evaluation Protocol and Reproducibility Controls}
All compared systems, adapters, task definitions, and official data were frozen before evaluation. The registered main schedule used the model label \code{DeepSeek-V4-Flash} and seeds \code{2026081901}, \code{2026081902}, and \code{2026081903}.

AlgoTune public search uses the first eight official train rows, a single editable task-local program, the original correctness verifier, and a 48-semantic-call ceiling. Timing runs in a fresh process with one warmup and three measured repetitions per public instance. Official replay uses the first 32 test rows for scheduled original endpoints.

\subsection{Evaluation Metrics and Comparison Conventions}
For a complete task $t$ and method $m$, let $q_{tms}$ be its frozen normalized evaluator score. The development task mean is $\bar q_{tm}=|S|^{-1}\sum_{s\in S}q_{tms}$, and the cross-task effect against baseline $b$ is $|T|^{-1}\sum_t(\bar q_{t,\epoch}-\bar q_{tb})$. The balanced development matrix has $|S|=3$, $|T|=12$. Official replay uses the analogous effect on the reported official-replay cohort. Development recovery and original replay do not share an identical endpoint population.

For heterogeneous AgentHPO metrics, let $d_t=1$ for higher-is-better metrics and $d_t=-1$ for lower-is-better metrics. Task-level means are
\begin{equation}
 \bar y_{tv}=\frac{1}{|S_{tv}|}\sum_{s\in S_{tv}}y_{tvs}.
\end{equation}
A paired effect uses the shared evaluation schedule $P_{tb}=S_{t,\epoch}\cap S_{tb}$:
\begin{equation}
 \delta_{tb}=\frac{1}{|P_{tb}|}\sum_{s\in P_{tb}}d_t(y_{t,\epoch,s}-y_{tbs}).
\end{equation}
The supplied paired report gives W/T/L and sign-test results separately from task-level means.

\section{Dataset Construction, Task Selection, and Data Splits}
\label{app:data-construction}
This section distinguishes four objects that are easy to conflate: a benchmark's native data, the task subset registered for this study, the public observations available during search, and the terminal observations reserved for endpoint evaluation. The supplied artifacts do not describe a newly crowdsourced dataset. Rather, they freeze task subsets and views from existing benchmark repositories. Table~\ref{tab:data-overview} gives the resulting analysis units.

\begin{table}[htbp]
\centering\small
\caption{Construction and split summary. ``Cell'' denotes one task--method--seed execution. Certificate cases are not treated as sampled benchmark rows.}
\label{tab:data-overview}
\begin{tabularx}{\linewidth}{@{}lYYY@{}}
\toprule
Collection & Registered population & Views or split & Analysis unit \\
\midrule
AlgoTune & 12 frozen official tasks; 5 methods; 3 seeds & First 8 official train rows for public search; first 32 official test rows for scheduled terminal replay & Task--method--seed cell; task-level mean \\
Math14 & 14 project-frozen objectives; 5 methods; 3-seed original schedule & Original schedule, fixed-seed terminal completion stage; no additional train/test partition documented & Task endpoint with stage provenance \\
AgentHPO & 6 reported environments; 5 EPOCH variants & Native task views & Paired task--variant observation \\
Discovery cases & 10 headline cases; 13 auxiliary groups (25 scoped results) & Search objects separated from exact replay, mutation controls, formal kernels, or independent implementations & Theorem, construction, finite universe, or executable certificate \\
\bottomrule
\end{tabularx}
\end{table}

\subsection{Task Cohort and Evaluation Data Separation}
The experimental cohort comprises the twelve-task schedule reported in Table~\ref{tab:data-overview}: affine transform, communicability, Convolve2D, zeta-zero count, cyclic independent set, real eigenvalues, FFT convolution, maximum common subgraph, NMF, Robertson ODE, real polynomial, and SVM. The supplied record identifies this set as a frozen benchmark subset.

For each task, search is restricted to the first eight rows of the official training data. These rows may influence code proposals, public correctness feedback, and endpoint selection. The first 32 rows of the official test data are reserved exclusively for scheduled replay of the already selected original endpoint; their outcomes are not fed back into that endpoint's search history. If a test failure prompts a new proposal, the resulting artifact constitutes a separate experiment and must not be reported as part of the original held-out replay. Row order and membership follow the frozen official-data revision documented in Table~S1; no reshuffling, stratification, or cross-validation is reported.

\subsection{Math14 Suite Construction and Staged Evaluation}

Math14 is a project-frozen suite of fourteen mathematical objectives. Its registered design forms a balanced factorial matrix of
$14\times5\times3=210$ cells, corresponding to fourteen objectives, five compared systems, and three registered seeds per objective--system pair. The suite was assembled and evaluated in stages: the objective list, system variants, seed schedule, and evaluation rules were fixed first; executable cells were then completed under the registered protocol; and the resulting records were frozen before aggregate analysis.

\subsection{AgentHPO Tasks and Native Evaluation Splits}
The reported AgentHPO ablation analysis covers the full EPOCH system and four component-removal variants targeting memory, the falsifier, proof admission, and the phase controller. The six reported environments span heterogeneous domains: game-playing reinforcement learning (ART 2048), natural-language inference (ModernBERT MNLI), noisy graph learning (NoisyGL Cora), semi-supervised facial-age regression (RankUp UTKFace), long-horizon time-series forecasting (SparseTSF ETTm1), and tabular regression (TabM California). Their target metrics are, respectively, win rate, classification accuracy, node-classification accuracy, mean absolute error (MAE), mean squared error (MSE), and root mean squared error (RMSE). Each environment's native optimization direction is preserved.

\subsection{How discovery-case evidence is constructed}
The mathematical and algorithmic cases do not use a statistical train/test split. Each begins with a frozen problem statement, comparator, editable object, and claim boundary. Search may inspect public algebraic identities, source constructions, or finite instances; admission then binds the selected object to exact verification obligations. Evidence is constructed in four noninterchangeable layers:
\begin{enumerate}
  \item \textbf{Semantic reconstruction:} rebuild the target independently from its definition (for example, tensor coordinates, AES truth tables, or selection ranks).
  \item \textbf{Exact or exhaustive replay:} check every coordinate or every member of the declared finite universe when the claim is finite.
  \item \textbf{Mutation controls:} corrupt a factor, coefficient, gate, recurrence term, or certificate field and require the verifier to reject it.
  \item \textbf{Formal or analytic bridge:} prove the reusable implication connecting the checked finite object to the stated theorem, while listing any remaining source-specific or representation-specific bridge.
\end{enumerate}
\begin{figure}[htbp]
\centering
\includegraphics[width=\linewidth]{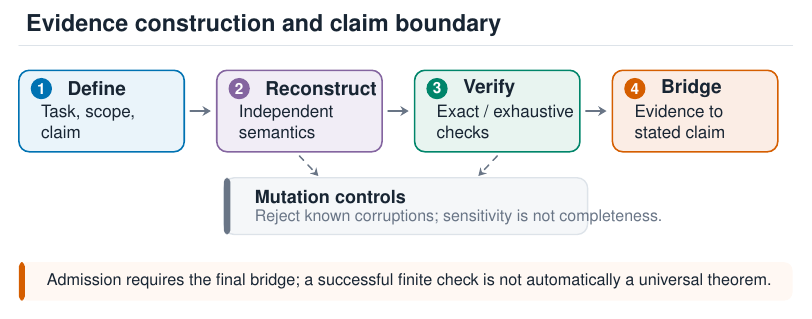}
\caption{Evidence construction from claim definition to the final theorem bridge. Exact replay establishes the declared finite object; mutation controls test sensitivity to selected errors; only the final bridge transfers those facts to the stated mathematical claim. The diagram is logical, not a count of experiments.}
\label{fig:s-evidence-pipeline}
\end{figure}
Discovery examples may guide the search, but the admitted statement is determined by the replay domain, not by how many exploratory examples happened to succeed. A finite exhaustive certificate and a sampled numerical check therefore receive different labels throughout Sections~\ref{app:algorithmic-cases}--\ref{app:auxiliary-results}.

\paragraph{Finite-certificate sufficiency.}
Let $U$ be the finite universe named by a claim, let $C$ be the admitted candidate, and let $P(C,u)$ be the semantic property to be established for every $u\in U$. Suppose an independent constructor enumerates $U$ without omission or duplication, and a sound exact checker returns true exactly when $P(C,u)$ holds. Then
\begin{equation}
 \left(|\widehat U|=|U|\ \wedge\ \widehat U=U\ \wedge\
 \bigwedge_{u\in\widehat U}P(C,u)\right)
 \Longrightarrow \forall u\in U,\ P(C,u).
\end{equation}
\begin{proof}
Take arbitrary $u\in U$. Completeness of the independent enumeration gives $u\in\widehat U$, and the accepted checker result for that element gives $P(C,u)$. Since $u$ was arbitrary, the universal finite claim follows. Soundness of the checker and equality $\widehat U=U$ are premises, not consequences of a large pass count; mutation controls can expose violations of soundness but cannot prove those premises by themselves.
\end{proof}

\subsection{Exclusions, deduplication, and contamination controls}
A task--method--seed key is counted once in a benchmark matrix. Retries caused by transport failure retain the same key and provenance; endpoint replays do not become new observations. Correct but non-improving proposals retain the endpoint selected by the frozen rule. Incorrect proposals are recorded as rejected trials, and an available verified fallback remains the method outcome when the contract provides one. In case studies, coordinate symmetries and isomorphism reductions may compress a search, but reported coverage is expanded back to the full declared universe before admission.

\section{Experimental Configuration and Execution Details}
\label{app:experimental-setup}
Table~\ref{tab:execution-config} consolidates the settings that are otherwise distributed across the supplied audit documents. 

\begin{table}[htbp]
\centering\small
\caption{Frozen execution configuration recoverable from the supplied artifacts.}
\label{tab:execution-config}
\begin{tabularx}{\linewidth}{@{}lY@{}}
\toprule
Field & Recorded configuration \\
\midrule
Compared methods & EPOCH, AI-Scientist v1, AI-Scientist v2, OpenEvolve, and AdaEvolve/SkyDiscover adapter \\
Model label & \code{DeepSeek-V4-Flash} for the main registered schedule \\
Random seeds & \code{2026081901}, \code{2026081902}, \code{2026081903} \\
AlgoTune proposal budget & 48 semantic calls in the main schedule\\
Editable artifact & One task-local program; interface and evaluator remain frozen \\
Public correctness & Original task verifier on the public view \\
Timing protocol & Fresh process; one warmup and three measured repetitions for each public instance \\
Terminal replay & First 32 official test rows; 600-second timeout per instance; exact selected endpoint only \\
AgentHPO variants & Full, no memory, no falsifier, no proof admission, no phase controller \\
Endpoint replacement & Candidate replaces the incumbent only when the applicable schema, scope, correctness, budget, challenge, and admission conditions pass \\
\bottomrule
\end{tabularx}
\end{table}

\subsection{Common execution lifecycle}
Every registered cell begins from the method-specific frozen revision and the same task contract. The runner records the initial artifact and verifier result before model interaction. At intervention $t$, the method receives only the public task view and the information permitted by its variant. A proposal must parse into the expected schema and stay within the editable surface. Candidate execution then records correctness, objective value, resource use, and any task-specific challenge. EPOCH additionally records evidence references, falsifier outcome, phase, and admission status. The event is appended even when rejected.

Endpoint replacement is separate from proposal evaluation. A higher public score cannot replace a verified incumbent if correctness or another required gate fails. At the budget boundary, the runner serializes the selected endpoint and its evidence root. Scheduled replay reconstructs precisely this serialization in a clean process. The replay result is linked to, but does not rewrite, the search record. This lifecycle also explains the ablations: each removal changes a declared part of the proposal or admission interface, rather than merely hiding a column during analysis.

\subsection{Evaluation Details for AlgoTune and Math14}

For each public AlgoTune instance, correctness is evaluated before timing. Timing is measured in a fresh process after one warmup run and aggregated across three measured repetitions according to the benchmark's frozen evaluator. Task-level means are computed by first averaging the three registered seeds, after which all tasks receive equal weight.

Math14 reports the project-normalized endpoint score returned by each objective-specific evaluator. Verification status is recorded separately from method failure: a failed method may still retain a task-defined fallback score.

\subsection{AgentHPO evaluation details}
AgentHPO preserves raw task units: win rate and accuracy are maximized, whereas MAE, MSE, and RMSE are minimized. The registered paired report orients each full-versus-removal difference so that positive favors full EPOCH, collapses to a task direction under its frozen tie rule, and applies an exact two-sided sign test across non-tied tasks.

Behavior diagnostics are computed from trace events rather than endpoint scores. Evidence-reference rate asks whether a decision cites a nonbaseline prior event; duplicate-configuration rate asks whether the emitted configuration exactly matches the baseline or an earlier configuration in the same cell; rejection rate records \code{admitted=false}. The alternative macro report aggregates within each task and then averages six tasks equally.

\section{Detailed Architecture and Evidence Protocol}
\label{app:architecture}
This section translates the main-text contract into protocol-level operations. The mathematical notation defines the quantities to preserve and verify, while the implementation details specify the concrete settings through which the architecture enforces its invariants.

\subsection{The research task as an executable contract}
A task is more than an objective function. It specifies an initial artifact, editable scope, allowed observations, correctness conditions, resource limits, and an endpoint rule. For a code task, the artifact may be a Python implementation of a function with a fixed input/output interface. For a tensor task, it may be a list of rank-one paths and a target-coordinate generator. For a mathematical statement, it includes the hypotheses and the meaning of any finite or formal evidence used to justify the conclusion. Without this surrounding contract, a high score is compatible with changing the problem rather than solving it.

The public view and terminal view have different roles. Public instances can influence proposal generation. A terminal view can evaluate the selected endpoint only under the declared protocol. If official test feedback is used to generate a new candidate, that candidate belongs to a new search experiment; it is not the original endpoint's replay. Similarly, a certificate verifier may expose a precise failed coordinate during search, whereas a held-out benchmark evaluation must not silently become an additional optimization loop.

A resource envelope has both accounting and attribution semantics. Candidate execution consumes task resources; proposal generation consumes semantic calls and associated model resources. A candidate that exceeds a valid registered time limit has failed within the method contract. An evaluator that cannot launch because a required device is unavailable creates a different state. The experiment must identify which boundary failed before deciding whether the cell has a score, or a fallback. This distinction is particularly consequential when comparing methods with different orchestration patterns.

\subsection{Evidence records and information exposed to generation}
An event should be interpretable independently of the model's narrative. It identifies the exact executable or symbolic candidate, evaluator version, and input version; metric and correctness fields record different outcomes; and attribution explains why the result can or cannot be used. An evidence reference points to this immutable event, not to a mutable natural-language statement such as ``the previous attempt worked.'' A subsequent claim can therefore be traced back to a specific observation without printing storage-level identifiers in the paper.

The bounded memory view is a projection of a longer evidence history. A projection can contain successful trials, rejected proposals, baseline observations, and relevant counterexamples without reproducing an entire transcript. This distinction matters because a context-window summary may omit details while the underlying record still retains them. The architecture's auditability claim concerns preservation of the full record and stable identifiers, not an assertion that the model always receives every historical event.

A useful proposal has at least four interpretable parts. Its hypothesis describes the mechanism intended to improve the objective. Its expected effect states the direction or observable change. Its evidence references identify earlier observations supporting the intervention. Its falsifier specifies a condition under which the mechanism should be rejected. For example, a faster convolution implementation may predict identical zero-padded linear convolution on rectangular shapes; a wraparound mismatch is a falsifier even when timing improves. An algebraic lift may predict a specified coefficient identity; a single mismatching semantic coordinate refutes that proposal.

Evidence can be useful without being positive. A tensor-chart obstruction excludes an entire local representation. A failure at a boundary input shows which invariant an optimization omitted. A repaired parser incident explains why a method appeared to produce no candidates. These observations should be stored with different types because they imply different next actions. Treating them all as low rewards loses both scientific information and the ability to diagnose the evaluation itself.

\subsection{Admission as a three-valued decision}

\begin{figure}[htbp]
\centering
\includegraphics[width=\linewidth]{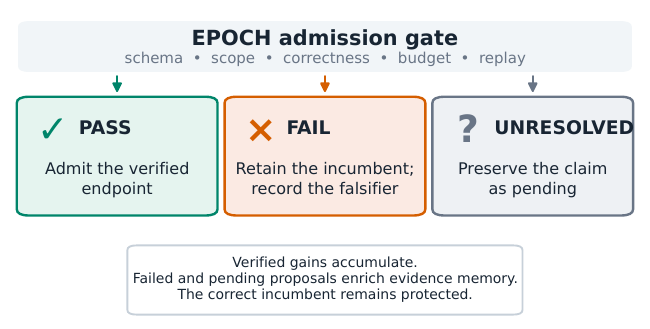}
\caption{EPOCH's three-valued admission gate. Passing proposals add verified gains; failed proposals retain the correct incumbent and contribute a falsifier; unresolved checks preserve the claim as pending. This separation lets EPOCH learn from adverse evidence while protecting an already verified endpoint.}
\label{fig:s-admission}
\end{figure}

Figure~\ref{fig:s-admission} makes the three outcomes explicit. Retention depends on the conjunction of the applicable gates, not a numerical average of their statuses.

For each applicable gate $g$, write $g(c)\in\{\mathsf{pass},\mathsf{fail},\mathsf{unresolved}\}$. Conjunction passes only if every required gate passes. Any demonstrated contract violation fails admission. An infrastructure incident that prevents a required check leaves admission unresolved; it does not establish either candidate correctness or candidate incorrectness. This gives an operational form to the main-text admission predicate.

The gates are logically heterogeneous. Parsing checks that the proposed artifact can be interpreted. Scope checks that changes remain within the task. Budget checks enforce the candidate's resource contract. Correctness checks the task semantics. Challenges test declared weak points or adverse cases. Replay checks reproducibility of the selected object.

A benchmark search endpoint and a theorem statement may require different final gates. A code endpoint can be admitted as correct on the registered validation instances without being admitted as universally correct on all inputs. A polynomial identity checked symbolically can support all values covered by its algebraic hypotheses. A finite certificate for 1,600 coordinates can support the whole finite tensor identity because those coordinates span the target object. The same number of sampled random matrices would have a different evidential meaning.

The word ``proof'' in proof admission therefore refers to the required justification, not to one proof technology. A Lean receipt can be a very strong terminal object, but only if the formal statement has been connected to the concrete candidate and task. Conversely, an exhaustive executable verifier can completely establish a finite property without a Lean translation.

\subsection{Protocol-level discovery loop}
The following pseudocode exposes the sequence of decisions without adding an unreported search heuristic.
\begin{quote}\small
\textbf{Input:} frozen contract, initial verified fallback when supplied, resource ledger, and empty append-only trial history.\\
\textbf{Initialize:} record the baseline, its verification receipt, and its resource use.\\
\textbf{While} an intervention remains allowed:\\
\hspace*{1em}Choose the current phase using the registered controller.\\
\hspace*{1em}Construct the bounded evidence view and request a proposal.\\
\hspace*{1em}Validate its schema, editable scope, and required evidence fields.\\
\hspace*{1em}Evaluate correctness, objective, and the applicable challenge.\\
\hspace*{1em}Attribute failures before deciding score eligibility.\\
\hspace*{1em}Append the candidate, observations, and gate reasons to history.\\
\hspace*{1em}Replace the endpoint only if admission and the comparison rule permit.\\
\textbf{Finalize:} reconstruct and replay the exact selected artifact.\\
\textbf{Seal:} bind the final claim, scope, contract, history, and replay receipts.
\end{quote}

The pseudocode deliberately separates evaluator observations from endpoint replacement. A candidate can be informative and still be ineligible for replacement.

\subsection{Endpoint safety and provenance invariants}
Suppose the baseline is verified under contract $\mathcal C$ and every endpoint replacement requires all applicable gates to pass. By induction on the number of interventions, the retained endpoint always has a passing receipt under that contract. At the base step this is the registered baseline. At a later step, a failed or unresolved candidate does not replace it, while an admitted replacement has its own passing receipt. This invariant explains why a misleading public score need not corrupt the endpoint.

The invariant is intentionally relative. If the correctness checker misses a class of inputs, a passing endpoint may still fail outside its registered coverage. If the checker itself is wrong, the invariant establishes protocol compliance rather than external truth. This is why the artifact must preserve the evaluator identity and why independent semantic reconstruction provides information beyond repeated execution of the same checker.

Append-only history gives a second invariant. A new event extends the history from $M_t$ to $(M_t,e_{t+1})$; it does not change the candidate or attribution of earlier events.

Claim sealing adds a binding invariant rather than a theorem of correctness. If the candidate, evaluator, inputs, or claim text changes, the archival record must create a distinct version. This versioning rule does not validate a theorem; it only ensures that a receipt cannot silently migrate to a different claim or artifact.

\subsection{Falsification, duplicates, and phase changes}
A repeated configuration can indicate useful replication, accidental repetition, or a failure to incorporate evidence. The AgentHPO trace metric counts equality with an earlier emitted configuration; it does not by itself distinguish those motivations. Consequently, a lower repetition rate is an operational signal, not a standalone measure of scientific efficiency. In the deterministic traps, the intended repeated behavior is specified by construction; in external tasks, endpoint outcomes remain necessary to interpret it.

A falsifier differs from an ordinary correctness test by being directed at the proposal's mechanism. A factor-sharing proposal must preserve ANF coverage and return ancillas; a certificate compression proposal must preserve all represented cases; a conjectured inequality must survive degenerate and equality regimes. The same challenge can discover a scientific failure and produce a useful input for the next proposal.

Phase changes organize how these observations affect search. Exploration seeks materially different hypotheses. Refinement invests in supported mechanisms. Finalization stops search and concentrates on replay and claim scope.

\subsection{Independent replay and trusted boundaries}
Replay independence has several levels. Process independence reconstructs the same candidate in a clean process. Implementation independence reconstructs the mathematical or executable target using another program. Representation independence uses a different computational model, such as multiplication-matrix norms versus resultants, or one irreducible finite-field model versus another. Formal reflection moves selected finite checks into a proof assistant. These levels can be complementary rather than interchangeable.

A strong replay package makes the interfaces explicit: exact input object, output statement, invariant checked, failure code, and resource boundary. For example, a QASM truth-table replay checks an emitted circuit; it does not verify the optimizer's claim of global minimality. A Lean determinant kernel proves an implication from a characteristic-polynomial hypothesis; it does not supply the representation-theoretic hypothesis. An exhaustive graph-catalogue pass verifies every listed record; it does not prove the catalogue contains every possible graph.

This layered view is also how the supplementary catalogue should be read. Each entry identifies the mathematical object and the strongest retained statement, then distinguishes construction, exhaustive finite checks, analytic arguments, and formal kernels. Results that are locally complete but globally unresolved remain useful evidence when their boundary is preserved.

\section{Engineering Endpoints Behind the AlgoTune Gains}
\label{app:engineering}
These cases connect benchmark gains to executable mechanisms. They are not counted among the four new algorithmic discovery candidates, because the core numerical and graph algorithms are established techniques. Their role is to show that high public scores correspond to recognizable task-solving strategies and to document the controls used to check them.

\subsection{Convolve2D: generic zero-padded real FFT}
The retained Convolve2D endpoint replaces the relevant direct computation with a generic zero-padded real FFT convolution. Zero padding must prevent circular aliasing, and the output slice must match the task's shape convention. A faster transform that returns a wrapped or shifted result would fail the semantic contract even if its timing is favorable.

The supplied audit records six shape controls and relative error below $9\times10^{-16}$. The public score moves from 0.5 to 1.0. These checks support the registered endpoint on the tested shape family and numerical tolerance; they do not make FFT convolution a new algorithm. AdaEvolve also later produced a clean generic solution, and the complete development table has a Convolve2D tie with AdaEvolve. The official-replay advantage against that baseline is correspondingly zero in the supplied task-effect table.

\subsection{Robertson ODE: selecting an appropriate stiff solver}
The Robertson endpoint uses LSODA with task-appropriate tolerances. The mechanism is solver and tolerance selection for the stiff system, rather than invention of a new integration method. Correctness depends on agreement with the task's solution accuracy and conservation behavior, not merely on returning a trajectory more quickly.

Six rate/time-horizon controls report mass error below $2\times10^{-12}$ and maximum error below $7.2\times10^{-8}$. The public score moves from 0.5 to 1.0.

\subsection{Maximum common subgraph: association graph and bitset clique search}
The retained implementation represents compatible vertex matches through an association graph and solves the corresponding clique problem with bitset operations. The correctness-critical interface is the correspondence between cliques and valid common-subgraph mappings: adjacency and injectivity restrictions must agree with the original task. Established association-graph and Bron--Kerbosch ideas supply the algorithmic basis.

The supplied independent checks include 145 optimum comparisons through size eight. The public score reaches 1.0 from a 0.5 baseline, and official replay retains a large positive task effect against every baseline. The evidence supports a successful implementation choice under the benchmark contract.

\subsection{NMF and zeta-zero count: scope of task-specific optimization}
NMF has development mean 0.70, while zeta-zero count has mean 0.68. Their evaluator-checked scores are reported at the task level and contribute to EPOCH's strong aggregate development result.

\section{AlgoTune Development and Official Replay}
\label{app:algotune}

\subsection{Development Coverage and Aggregate Effects}

Table~\ref{tab:s-development-coverage} summarizes the complete development matrix. Each method contributes 36 evaluated cells, corresponding to twelve tasks and three registered seeds. EPOCH achieves the highest mean normalized score, 0.65, compared with 0.53 for the strongest baseline.

\begin{table}[htbp]
\centering
\small
\caption{Complete descriptive development coverage.}
\label{tab:s-development-coverage}
\begin{tabular}{@{}lrr@{}}
\toprule
Method & Cells & Mean \\
\midrule
\epoch{} & 36 & 0.65 \\
AdaEvolve & 36 & 0.53 \\
AI-Scientist v1 & 36 & 0.52 \\
OpenEvolve & 36 & 0.50 \\
AI-Scientist v2 & 36 & 0.50 \\
\bottomrule
\end{tabular}
\end{table}

Table~\ref{tab:s-development-effects} distinguishes the supplied original reported effects from effects computed over the complete development matrix. EPOCH has a positive mean effect against every baseline under both summaries. The win/tie/loss profiles further indicate that the aggregate advantage varies across tasks rather than arising from a uniform improvement.

\begin{table}[htbp]
\centering
\small
\caption{Distinct AlgoTune development summaries. Original reported effects are supplied aggregates from a separate cohort, whereas complete development effects and win/tie/loss counts are computed over the complete development matrix.}
\label{tab:s-development-effects}
\begin{tabular}{@{}lrrc@{}}
\toprule
Baseline & Original reported effect & Complete development effect & Complete W/T/L \\
\midrule
AI-Scientist v1 & +0.17 & +0.13 & 6/0/6 \\
AI-Scientist v2 & +0.20 & +0.15 & 6/0/6 \\
OpenEvolve & +0.19 & +0.15 & 10/0/2 \\
AdaEvolve & +0.16 & +0.12 & 8/1/3 \\
\bottomrule
\end{tabular}
\end{table}

\subsection{Official-Replay Task Effects}

\begin{figure}[htbp]
\centering
\includegraphics[width=\linewidth]{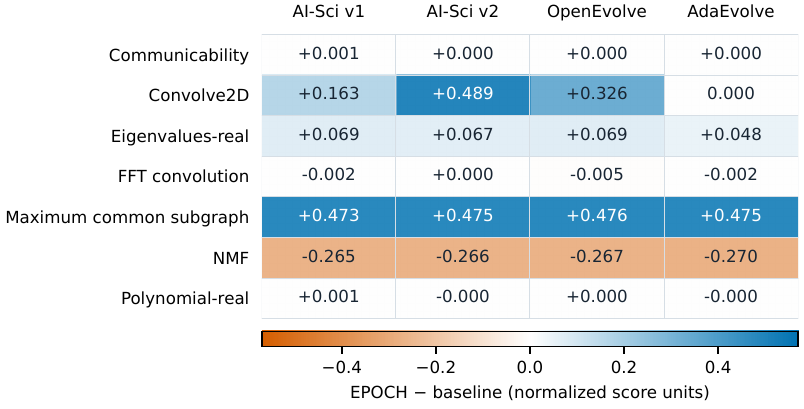}
\caption{Reported official-replay task effects using the original frozen endpoints. Each value is the score of \epoch{} minus that of the corresponding baseline, rounded to three decimal places; signed zeros preserve the direction of effects that round to zero. The largest gains occur on maximum common subgraph and Convolve2D, while NMF exhibits a consistent reversal.}
\label{fig:s-replay}
\end{figure}

Figure~\ref{fig:s-replay} presents the reported task-level official-replay effects. EPOCH achieves its largest transferable gains on maximum common subgraph and Convolve2D, together with consistent positive effects on eigenvalue computation. These gains outweigh the reversal on NMF, yielding a positive mean replay effect against every baseline. The resulting profile reflects substantial task-specific improvements rather than a uniform shift across all tasks.

\section{Math14 Evaluation and Transfer}
\label{app:math14}

Math14 is a project-frozen suite of fourteen mathematical objectives evaluated across five methods and three registered seeds, yielding a balanced design of $14\times5\times3=210$ cells. Each objective reports a project-normalized endpoint score through its task-specific evaluator.

\begin{table}[htbp]
\centering
\small
\caption{Aggregate Math14 results.}
\begin{tabular}{@{}lrr@{}}
\toprule
Method & Mean & \epoch{} advantage \\
\midrule
\epoch{} & 0.57 & --- \\
OpenEvolve & 0.56 & +0.01 \\
AdaEvolve & 0.53 & +0.04 \\
AI-Sci v2 & 0.52 & +0.05 \\
AI-Sci v1 & 0.51 & +0.06 \\
\bottomrule
\end{tabular}
\end{table}

EPOCH achieves the highest aggregate Math14 score, with a mean of 0.57. Its pairwise win/tie/loss records are 7/4/3 against AI-Scientist v1, 7/5/2 against AI-Scientist v2, 4/3/7 against OpenEvolve, and 5/3/6 against AdaEvolve. Although the task-level comparisons are mixed, EPOCH's larger gains on several objectives produce the strongest overall mean. The discovery portfolio below presents the exact constructions, counterexamples, reductions, and proofs underlying documented outcomes.

\section{AgentHPO Ablation and Mechanism Behavior}
\label{app:agenthpo}
\label{app:ablation}

\subsection{Ablation Results}

The reported AgentHPO analysis covers six tasks across the full system and its four component-removal variants.

\begin{table}[htbp]
\centering
\small
\caption{Task-level means on the native task metrics. Full, M, F, P, and C denote full \epoch{}, no memory, no falsifier, no proof admission, and no phase controller, respectively. MAE, MSE, and RMSE denote mean absolute error, mean squared error, and root mean squared error.}
\label{tab:rawhpo}
\setlength{\tabcolsep}{3.5pt}
\begin{tabular}{@{}llrrrrr@{}}
\toprule
Task & Metric & Full & M & F & P & C \\
\midrule
ART 2048 & Win rate $\uparrow$ & 0.07 & 0.15 & 0.03 & 0.03 & 0.03 \\
ModernBERT MNLI & Accuracy $\uparrow$ & 47.72 & 49.79 & 47.46 & 44.69 & 48.13 \\
NoisyGL Cora & Accuracy $\uparrow$ & 0.71 & 0.70 & 0.70 & 0.70 & 0.70 \\
RankUp UTKFace & MAE $\downarrow$ & 17.13 & 21.44 & 30.56 & 19.20 & 27.60 \\
SparseTSF ETTm1 & MSE $\downarrow$ & 0.33 & 0.33 & 0.33 & 0.33 & 0.34 \\
TabM California & RMSE $\downarrow$ & 0.52 & 0.54 & 0.55 & 0.53 & 0.54 \\
\bottomrule
\end{tabular}
\end{table}

\begin{table}[htbp]
\centering
\small
\caption{Paired task-level comparisons between the full system and each component-removal variant. W/T/L denotes the number of task-level wins, ties, and losses for full \epoch{}.}
\label{tab:agenthpo-paired}
\begin{tabular}{@{}lrc@{}}
\toprule
Removed component & Tasks & Full W/T/L \\
\midrule
Memory & 6 & 4/0/2 \\
Falsifier & 6 & 4/1/1 \\
Proof admission & 6 & 6/0/0 \\
Phase controller & 6 & 5/0/1 \\
\bottomrule
\end{tabular}
\end{table}

The full system outperforms the no-proof-admission variant on all six tasks and the no-phase-controller variant on five of six tasks. It also records four task-level wins against both the no-memory and no-falsifier variants. These comparisons identify proof admission and phase control as the most consistently beneficial components, while the effects of memory and falsification vary more strongly across environments.

\subsection{Observed Mechanism Behavior}

A prior-evidence citation is a decision whose \code{evidence_refs} includes an identifier greater than zero, with identifier zero reserved for the baseline observation. A repeated configuration is a \code{new_config} identical to the baseline or to an earlier configuration within the same task-seed run. Admission rejection corresponds to \code{admitted=false}.

\begin{table}[htbp]
\centering
\small
\caption{Matched mechanism-behavior comparisons. Differences are reported in percentage points (pp), and favorable tasks indicate agreement with the intended behavior of the corresponding component.}
\label{tab:agenthpo-behavior}
\begin{tabularx}{\linewidth}{@{}Yrrrr@{}}
\toprule
Comparison & Full & Removal & Difference (pp) & Favorable tasks \\
\midrule
Prior evidence cited, vs no memory & 80.00\% & 42.50\% & +37.50 & 5/6 \\
Repeated configurations, vs no falsifier & 34.29\% & 50.00\% & -15.71 & 3/6 (2 ties) \\
Repeated configurations, vs no proof & 32.00\% & 48.00\% & -16.00 & 4/6 (1 tie) \\
Repeated configurations, vs no phase & 33.75\% & 43.75\% & -10.00 & 3/6 (1 tie) \\
Admission rejection, vs no proof & 34.67\% & 0.00\% & +34.67 & 6/6 \\
\bottomrule
\end{tabularx}
\end{table}

Memory increases the use of prior evidence by 37.50 percentage points. The falsifier, proof-admission module, and phase controller each reduce repeated configurations, while proof admission introduces selective rejection across all six tasks. Together, these patterns show that the components influence both the information used to generate proposals and the mechanisms governing their acceptance and progression.

\clearpage
\section{Four Algorithmic Discovery Cases}
\label{app:algorithmic-cases}
\label{app:algorithms}
The following are documented research outcomes with heterogeneous budgets and verification layers. Border rank is denoted $\brank$, ordinary tensor rank $R$, and Waring rank $R_{\mathrm W}$. General constructions are stated only under the field, invertibility, and nonzero-denominator conditions required by their certificates. External record priority remains a separate question.

Each proof is organized as a chain rather than a single undifferentiated ``verified'' label. The following subsections give the equations, constructions, exhaustive counts, corruption tests, and unresolved links. Their common coverage and remaining bridges are summarized together with the mathematical cases in Table~\ref{tab:s-proof-gaps}.

\begin{table}[htbp]
\centering\scriptsize
\caption{Cross-case evidence coverage and remaining theorem bridges.}
\label{tab:s-proof-gaps}
\setlength{\tabcolsep}{3pt}
\begin{tabularx}{\linewidth}{@{}p{0.18\linewidth}*{5}{>{\centering\arraybackslash}X}@{}}
\toprule
Case & Statement & Algebraic reduction & Finite/exact verification & Formal kernel & Final bridge \\
\midrule
Reverse transfer & In scope & In scope & In scope & In scope & Partial \\
Hyperfactor & In scope & In scope & In scope & Partial & Partial \\
Big-centroid & In scope & In scope & In scope & Partial & Partial \\
Median-23 & In scope & In scope & In scope & Partial & Open \\
Hypercube & In scope & In scope & In scope & In scope & Partial \\
$H_\beta$ & In scope & In scope & In scope & In scope & Partial \\
Three-state $K$ & In scope & In scope & In scope & Partial & Partial \\
Full Perazzo & In scope & In scope & In scope & Partial & Partial \\
Heisenberg & In scope & In scope & In scope & Partial & Open \\
Online embedding & In scope & In scope & In scope & Partial & Open \\
\bottomrule
\end{tabularx}
\end{table}

\subsection{Reverse 2-adic transfer and obstruction-guided Hasse lifting}

\begin{figure}[htbp]
\centering
\includegraphics[width=\linewidth]{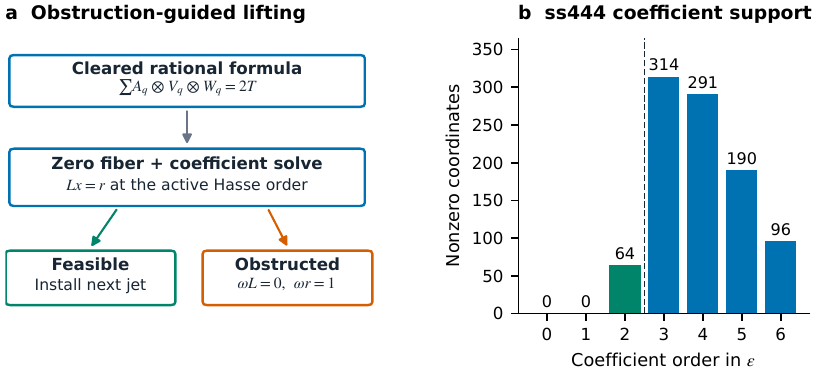}
\caption{Reverse transfer as an evidence-guided construction. (a) At each active Hasse order, an exact linear solve either installs a jet or returns a checkable cokernel obstruction. A changed chart may then yield a different solve. (b) The supplied ss444 certificate has coefficient support weights $(0,0,64,314,291,190,96)$. The dashed line separates the verified target coefficient at order two from higher-order terms.}
\label{fig:s-transfer}
\end{figure}

Figure~\ref{fig:s-transfer} links the algebraic control loop to the concrete coefficient profile.

\paragraph{Problem and mechanism.}
A rational rank-$r$ tensor decomposition can fail to reduce directly modulo two because its factors contain powers of two in their denominators. Suppose clearing the first-mode denominators gives integral factors satisfying
\begin{equation}
 \sum_{q=1}^r A_q\otimes V_q\otimes W_q=2T.
\end{equation}
Reduction modulo two yields a zero-fiber point $p=((a_q,b_q,c_q))$ of the Segre-sum parametrization. Its differential is
\begin{equation}
 d\Phi_p(\dot a,\dot b,\dot c)=\sum_q
 (\dot a_q\otimes b_q\otimes c_q+a_q\otimes\dot b_q\otimes c_q
 +a_q\otimes b_q\otimes\dot c_q).
\end{equation}
If $T\bmod2$ lies in this image, the factor paths satisfy
\begin{equation}
 \sum_q(a_q+\varepsilon\dot a_q)\otimes(b_q+\varepsilon\dot b_q)
 \otimes(c_q+\varepsilon\dot c_q)=\varepsilon T\pmod{\varepsilon^2}.
\end{equation}
Expanding the product proves the assertion and gives $\brank_{\overline{\mathbb F}_2}(T)\leq r$. A mechanized kernel proves the first-order expansion over a commutative ring. The arithmetic reduction and concrete tensor binding remain explicit obligations outside the generic identity.

\paragraph{Higher-order search.}
Different valuation strata require Hasse jets. The algorithm installs lower-order coefficients, builds the remaining linear tangent map, and tests the residual for membership in its image. Failure produces a cokernel functional $\omega$ with $\omega(\mathrm{residual})=1$, which constrains a change of chart or support. All retained coefficients are replayed exactly. This converts an unsuccessful lifting attempt into a reusable algebraic obstruction.

\begin{quote}\small
Clear denominators and assign each term its activation height. For each coefficient order, subtract contributions already fixed by lower jets. Solve the exact linear residual equation when feasible; otherwise retain a separating cokernel witness and revise the chart. After completing the jets, replay every coefficient and tensor coordinate before admission.
\end{quote}

\paragraph{Flagship certificate.}
For the $10\times10\times16$ structure tensor of multiplying two symmetric $4\times4$ matrices, 30 rank-one polynomial paths satisfy
\begin{equation}
 \sum_{q=1}^{30}A_q(\varepsilon)\otimes B_q(\varepsilon)\otimes C_q(\varepsilon)
 =\varepsilon^2T_{\mathrm{ss444}}\pmod{\varepsilon^3}.
\end{equation}
The valuation strata have sizes 4/13/13. The first-order map has rank 68 and the second-order system rank 328. The known second-order nonlinear contribution has weight 234. Truncated coefficient weights are $(0,0,64)$; full polynomial weights are $(0,0,64,314,291,190,96)$. The target has 64 nonzero coordinates among 1,600 ambient coordinates. The mechanized certificate checks all ambient coordinates and the semantic target, and rejects corrupted factor data.

Across 86 scanned rational formulas, 43 are directly applicable and all 43 transfer, spanning 24 coordinate-distinct targets and 45,242 independently replayed coordinates. Eleven strict border improvements are reported internally, including $\brank(T_{\mathrm{st444}})\leq20$. The denominator is applicable formulas, not all scanned formulas. Bini interpolation gives the retained scalar-count bound
\begin{equation}
 R(T_{\mathrm{ss444}}^{\otimes N})\leq(4N+1)30^N.
\end{equation}
It first beats $32^N$ at $N=92$ in that model; the retained extension-field descent model gives a crossover at $N=170$. Neither number is a runtime benchmark.

\subsubsection{Coefficient-level derivation and obstruction certificates}
For clarity, write a trilinear factor path as $a(\varepsilon)=a_0+\varepsilon a_1+\varepsilon^2a_2+\cdots$, and similarly for $b,c$. The coefficient of order $k$ in a rank-one path is
\begin{equation}
 [\varepsilon^k](a\otimes b\otimes c)=\sum_{i+j+\ell=k}a_i\otimes b_j\otimes c_\ell.
\end{equation}
At order one, exactly three terms occur, giving the Segre differential. At order two, the terms involving a new second jet are linear in that jet, while products of first jets are already known. This is why each lifting stage separates into a known nonlinear contribution and an exact linear residual problem. The decomposition is structural, not an approximation to the coefficient equations.

Suppose a stage asks for $Lx=r$ over a finite field. If no solution exists, row reduction can produce a functional $\omega$ satisfying $\omega L=0$ but $\omega r\ne0$; rescale to $\omega r=1$ when applicable. Anyone can verify this witness by multiplication. It proves that the current residual is outside the active tangent image. It does not prove that the tensor has no decomposition of the desired rank, since a new base point, chart, support, or higher-order ansatz changes $L$ and possibly $r$.

This distinction is central to the search loop. A witness excludes a precise local proposal family and can constrain the next representation. It cannot be admitted as a global tensor-rank lower bound. Repeatedly increasing a solver timeout would not resolve an exact cokernel obstruction within the same chart; changing the chart is a mathematically different intervention.

\subsubsection{From a truncated identity to a border upper bound}
Let the full polynomial rank-one sum be $F(\varepsilon)$ and suppose coefficient verification establishes
\begin{equation}
 F(\varepsilon)=\varepsilon^hT+\varepsilon^{h+1}E_1+\cdots+\varepsilon^{h+e}E_e.
\end{equation}
For nonzero $\varepsilon$, the rescaled tensor $\varepsilon^{-h}F(\varepsilon)$ has rank at most the number of paths, because the scalar can be absorbed into one factor of each term. As $\varepsilon$ tends formally or algebraically to zero, its limit is $T$. This supplies the border upper bound over the stated closure. It says nothing about whether $T$ itself has a rank decomposition of that same length without a limiting parameter.

For a tensor power $N$, the error degree is at most $eN$. Interpolating at $eN+1$ suitable field points recovers the constant coefficient of the powered rescaled expression, giving the retained $(eN+1)r^N$ exactification pattern. Availability of interpolation points and extension-field descent costs must be accounted for separately. The ss444 result uses $e=4$, yielding $(4N+1)30^N$; the later tail-elimination cases improve $e$ without necessarily changing $r$.

\subsubsection{Semantic reconstruction and negative controls}
A certificate must be checked against the intended matrix operation, not only against a tensor imported from the same optimizer. The independent route reconstructs the operation tensor from input/output matrix entries, decodes the official rational factors, and checks every coefficient against those semantics. This separates factor parsing, target construction, and polynomial expansion as possible error sources.

The ss444 coefficient weights are useful compact diagnostics, but weights alone cannot prove identity: two different tensors can have the same number of nonzero coordinates. Admission therefore uses the full 1,600-coordinate comparison, with the weights serving only as a support-count summary. The one-bit corruption test demonstrates sensitivity to a chosen factor error. It complements, rather than replaces, the complete coordinate proof.

\subsection{Hyperfactor and cross-output AES sharing}

\begin{figure}[htbp]
\centering
\includegraphics[width=\linewidth]{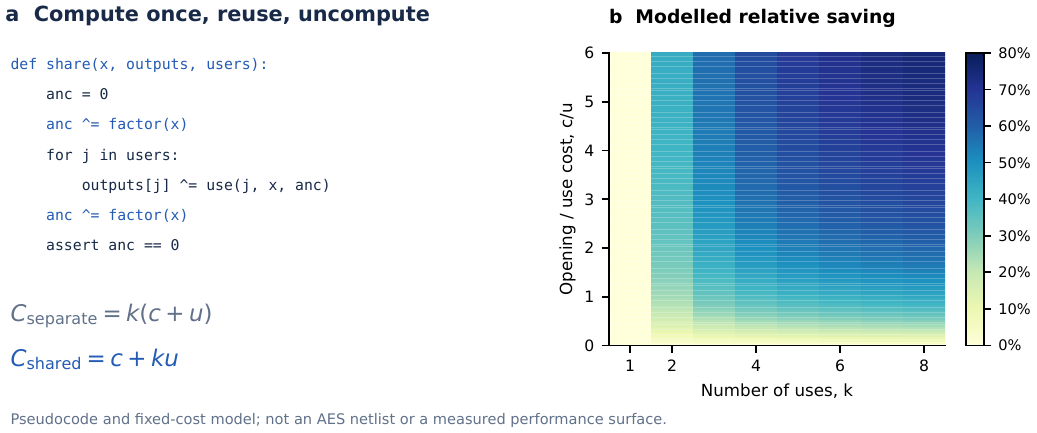}
\caption{Factor-sharing mechanism and an illustrative cost model. The pseudocode keeps $x$ unchanged, computes a factor into a zero ancilla, reuses it, and uncomputes it. If compute/uncompute costs $c$ and each use costs $u>0$, separate uses cost $k(c+u)$ and shared uses cost $c+ku$. The plotted relative saving is $(k-1)c/[k(c+u)]$, for integer $k$. This simplified surface is not fitted to AES measurements and ignores allocation, routing, and factor-dependent use costs; it is not the emitted circuit.}
\label{fig:s-sharing}
\end{figure}

Figure~\ref{fig:s-sharing} explains why cross-output factor selection can change compute/uncompute cost. Its model isolates reuse count and the opening-to-use cost ratio; the recorded AES reduction from 10,746 to 9,234 T gates is a separate matched measurement, not a point calibrated on that surface. The identity, schedule, and complete circuit each require their own checks.

\paragraph{Optimization problem.}
Let $G$ be residual groups and $F$ the frozen factor library. Opening factor $f$ costs $c_f$; assigning residual $g$ to it costs $r_{gf}$. Solve
\begin{equation}
 \min_{x,y}\sum_f c_fy_f+\sum_{g,f}r_{gf}x_{gf},\qquad
 \sum_f x_{gf}=1,\quad x_{gf}\leq y_f,\quad x_{gf},y_f\in\{0,1\}.
\end{equation}
A selected factor is computed into an ancilla, reused for assigned outputs, and uncomputed. The Boolean kernel is $(a\wedge h)\oplus(b\wedge h)=(a\oplus b)\wedge h$, mechanized together with its output-family lifting. Circuit equivalence requires additional execution checks.

\begin{table}[htbp]
\centering\small
\caption{AES T-counts under the frozen representation and cost model. The last two rows share the same PyZX post-processing pass.}
\begin{tabular}{@{}lr@{}}
\toprule
Stage & T-count \\
\midrule
Exact pair baseline & 25,886 \\
Eight independently optimized outputs & 12,362 \\
Cross-output factor merge & 10,696 \\
Matched independent arm plus PyZX & 10,746 \\
Matched shared arm plus PyZX & 9,234 \\
\bottomrule
\end{tabular}
\end{table}

The construction covers 1,013 ANF occurrences, shares 37 repeated nonzero-cost factors, and uses 1,528 CCX gates and 23 qubits in the reported pre-PyZX circuit. HiGHS and an independent SCIP implementation agree on the fixed-charge objective. An independent verifier reconstructs AES inversion and affine-output ANFs, checks exact monomial coverage, and simulates all 256 inputs through QASM while checking ancilla restoration. Removing a CCX is a negative control. Matched post-processing saves 1,512 T gates; the comparison against the frozen pair baseline saves 15,190 T gates.

The independent optimization replay, ANF reconstruction, and circuit receipt agree on the registered construction. Optimality applies only to the registered factor library and cost model. Global AES circuit optimality, routing, depth, and noisy hardware performance are not established.

\subsubsection{Why arbitrary-arity sharing differs from pair matching}
A pairwise optimizer considers sharing between two residual expressions at a time. A factor that benefits three or more groups can be unprofitable in every locally selected pair and still profitable when its opening cost is paid only once for the whole group. The binary variable $y_f$ represents this global activation decision, while the $x_{gf}$ variables assign residual use. Their coupling captures a cost that cannot be represented by independent residual choices alone.

For a factor used in $k\geq2$ outputs, let $c>0$ be its registered compute--uncompute cost and $u>0$ the per-output use cost. Separate realization costs $k(c+u)$, whereas a legal shared schedule costs $c+ku$. Hence
\begin{equation}
 k(c+u)-(c+ku)=(k-1)c>0,
 \qquad
 \frac{k(c+u)-(c+ku)}{k(c+u)}=\frac{(k-1)c}{k(c+u)}.
\end{equation}
This establishes the saving claimed by the schematic whenever the same factor remains live across all uses and the final inverse computation restores the ancilla. It does not establish that a factor is semantically common to those outputs or that routing preserves the same cost; those are separate certificate obligations.

The optimization model also distinguishes logical sharing from reversible execution. A classical expression DAG may reuse a value without specifying how temporary storage is restored. A reversible circuit must include the computation of the factor, the uses that leave it available, and the inverse computation returning the temporary register to its initial value. The opening cost $c_f$ therefore includes the registered compute/uncompute expense. A model that omits the inverse computation would optimize the wrong cost contract.

\subsubsection{Constructing a circuit from a selected assignment}
For each opened factor, construct its value on a designated ancilla using the registered factor implementation. Apply the residual-specific controlled operations to the assigned outputs in the scheduled order. Because the factor register is used as a control and its input data remain available under the contract, run the inverse factor computation to reset that ancilla. The emitted schedule is then translated to QASM and checked as a complete circuit.

This procedure requires more than an optimal integer-program objective. The assignment must correspond to the intended ANF occurrences; every occurrence must be accounted for under the representation's cancellation convention; control dependencies must be legal; and uncomputation must preserve outputs. A solver can be correct while the schedule generator is wrong. The independent ANF and QASM layers address these different interfaces.

The Boolean identity follows by distributivity in the two-element Boolean ring. For bits, XOR is addition and AND is multiplication, so $(a+b)h=ah+bh$. Applying the identity at each output index justifies shared algebraic factors. The formal identity does not encode the entire ancilla schedule or a decomposition of every CCX into T gates, which is why the full-circuit and cost checks remain separate.

\subsubsection{Interpreting the cost comparisons}
The first comparison, 25,886 to 10,696 T gates, measures the combined effect of moving from the frozen pair baseline to the shared construction. The intermediate 12,362 value reveals an additional reduction of 1,666 T gates from cross-output merging after independent-output Hyperfactor optimization. The matched PyZX comparison isolates a shared-versus-independent contrast under the same post-processing pass: 10,746 versus 9,234, a difference of 1,512.

These numbers should not be combined as if they were measurements of the same single intervention. The matched post-processing comparison provides the cleanest evidence that sharing remains advantageous after that specific optimizer. The larger pair-baseline difference reflects a broader change of construction. Neither establishes minimal T count among all AES circuits, because the factor library, ancilla budget, synthesis basis, and permitted schedules delimit the search space.

Finally, enumerating all 256 inputs is exhaustive for the registered eight-bit input contract. It checks the implemented truth table and ancilla restoration under that contract. It does not directly measure routing overhead, depth under a physical architecture, or behavior under noise. The contribution is a verifiable circuit construction and a cost comparison in a fixed model.

\subsection{All-parameter Big-centroid constructor}
For $d\geq3$ and ordered dimensions $u_1\leq\cdots\leq u_d$, set $q=\prod_{s=2}^d u_s$. Choose full-column-rank matrices $V^{(s)}\in\mathbb F^{u_s\times u_1}$ with nonzero entries and left inverses $L^{(s)}V^{(s)}=I$. This existence condition is part of the construction contract; a finite set of field tests does not prove existence over every small field.

For $J=(j_2,\ldots,j_d)$, define first-mode coordinates $a_J(t)=\prod_{s=2}^d V^{(s)}_{j_s,t}$. The perturbation for mode $i\geq2$ uses
\begin{equation}
 \frac{L^{(i)}_{t,j_i}}{\prod_{s\geq2,\,s\ne i}V^{(s)}_{j_s,t}}.
\end{equation}
Summing over $j_i$ yields $\sum_j L^{(i)}_{t,j}V^{(i)}_{j,t'}=\delta_{tt'}$. Thus target linear coordinates appear with the required coefficients and unwanted coordinates cancel. Subtracting $u_1$ constant correction tensors yields a $q+u_1$-term first-order degeneration. The retained flattening lower bound matches $q+u_1$ in the stated Big-centroid family.

The generic mechanized first-order kernel proves constant cancellation and the tangent coefficient identity. The constructor separately checks matrix inverses, denominators, and coordinates. The supplied audit records 390 parameter/field controls and 278,904 exact identities, including odd prime fields and $\mathrm{GF}(2^k)$ for tested $k\geq2$. The $(2,2,2,3)$ example over $\mathbb F_5$ has 14 terms, 30,184 ambient coordinates, and degree-0--4 support weights $(0,44,288,240,64)$.

The result is a symbolic constructor with an independent exact verifier under the stated assumptions, not an improvement in the matrix-multiplication exponent. Related tensor-centroid theory supplies context \citep{centroidsource}.

\subsubsection{The left-inverse identity as a certificate generator}
The constructor's core operation is local cancellation. Fix mode $i$ and all indices except $j_i$. Multiplying the perturbation coefficient by the base-coordinate factors cancels every denominator from the modes held fixed. The remaining sum over $j_i$ is exactly an entry of $L^{(i)}V^{(i)}$. It equals one when the two first-mode labels coincide and zero otherwise. Thus the same small-matrix identity both selects desired coordinates and annihilates undesired ones.

This is more informative than solving the full ambient tangent system independently for each dimension tuple. The latter can certify one instance but hides why the instance works. The left-inverse construction exposes a dimension-parametric algebraic mechanism and emits witnesses whose validation reduces to matrix products, nonzero-denominator checks, and coordinate expansion. Its usefulness is not tied to one unusually favorable numerical seed.

The $q$ base paths contribute a constant term. The retained construction expresses that constant as $u_1$ correction tensors, so adding their negatives produces zero constant coefficient. The prescribed first-order perturbations then yield the target. This is the reason for the $q+u_1$ term count; it is not inferred from a numerical fit. The matching flattening argument is needed to call the bound minimal within the stated family.

Concretely, let $\mathrm{Flat}(T)$ be the registered linear flattening of the Big-centroid tensor. A rank-one tensor maps to a matrix of rank at most one, so subadditivity gives $\operatorname{rank}\mathrm{Flat}(T')\leq r$ for every tensor $T'$ of rank at most $r$. Matrix rank is upper semicontinuous under specialization; hence the same inequality holds for every tensor of border rank at most $r$. Once the displayed block structure is shown to have
\begin{equation}
 \operatorname{rank}\mathrm{Flat}(T)=q+u_1,
\end{equation}
it follows that $\brank(T)\geq q+u_1$. The first-order construction gives the reverse inequality, and therefore equality. The remaining obligation is to prove that block-rank identity uniformly under the stated field hypotheses; checking finitely many parameter tuples alone would not suffice.

\subsubsection{Field assumptions and small-field controls}
A formula containing division is meaningful only when its denominators are nonzero. The constructor therefore supplies the actual coefficient matrices and their left inverses rather than merely stating that suitable matrices should exist. In finite fields, the matrix-existence step needs a field-aware construction. A characteristic-zero interpolation formula cannot automatically be reduced modulo a prime that collapses distinct parameters or makes denominators vanish.

The supplied source claims a uniform field construction over odd prime fields and binary extension fields of degree at least two. Its 390 controls and 278,904 identities provide a finite audit of the implementation. The generic first-order Lean kernel does not formalize this full existence theorem. In this manuscript the explicit left-inverse hypotheses are retained in the displayed construction so that the general algebraic implication is not confused with the separate existence and instantiation layer.

The example's 30,184 ambient coordinates illustrate another distinction: a sparse target can require checking many zero coordinates. Support weights $(0,44,288,240,64)$ summarize the polynomial path by degree, but complete verification must check where those coefficients occur. A correct target weight with an incorrect coordinate assignment would not establish the intended tensor degeneration.

\subsubsection{Algorithmic and mathematical roles of the result}
As an algorithmic result, Big-centroid is a deterministic generator producing factor paths and a checkable payload from a dimension tuple and an admitted field. As a mathematical result, its correctness is a symbolic implication from the matrix identities, constant cancellation, and flattening lower bound. These roles are related but not identical: a general proof without an implementation does not provide an executable artifact, while finite implementation tests without the general argument do not establish all parameters.

The symmetric Full Perazzo case developed later uses a different interpolation organization and has additional Waring-rank and geometric conclusions. We keep the two case identities distinct to avoid counting one formula twice as unrelated evidence. They demonstrate a shared transition from pattern observation to explicit construction, but their targets, assumptions, and lower-bound arguments differ.

\subsection{Median-of-23 in 54 comparisons}
Let $V_i(n)$ denote the minimum worst-case comparisons needed to select the $i$th smallest of $n$ distinct totally ordered elements. The retained candidate is $V_{12}(23)\leq54$. The supplied comparison point is a 55-comparison Hadian--Sobel construction; external selection-literature review remains separate from the certificate \citep{selection}.

An early local candidate was falsified by paths requiring 56 comparisons. The retained construction freezes those failures, classifies the over-budget frontier by poset structure, and synthesizes budget-specific continuations. The exact path decomposition is
\begin{equation}
1{,}439{,}190{,}016=1{,}427{,}981{,}824+11{,}208{,}192.
\end{equation}
The first summand already fits budget 54. The remaining occurrences form 1,454 classes:
\begin{table}[htbp]
\centering\small
\caption{Repaired median frontier. Every base depth plus continuation budget equals 54.}
\begin{tabular}{@{}rrrr@{}}
\toprule
Base depth & Classes & Occurrences & Continuation budget \\
\midrule
32 & 15 & 47,104 & 22 \\
33 & 422 & 2,850,304 & 21 \\
34 & 1,017 & 8,310,784 & 20 \\
\midrule
Total & 1,454 & 11,208,192 & --- \\
\bottomrule
\end{tabular}
\end{table}

The dispatcher runs the frozen median-17 base strategy, inserts six new items relative to the selected median, and invokes a seven-item prefix policy. If three new items lie below the selected item, it returns that item. Otherwise it selects the relevant side, constructs the residual poset, removes elements certified outside the target rank, canonically relabels the state, and executes a budget-decreasing residual DAG. Sixteen policies cover the classes.

The reported Lean layers cover transitive bitmask insertion, rank semantics of leaves, rank-preserving removal/reduction, bijective relabeling, preservation of distinguished index zero, and well-founded policy termination. They also check 4,100 prefix nodes, 9,200 residual nodes, 84,894 transfers, class counts, and global path arithmetic. Negative controls corrupt rank decrements, relabeling, residual ranks, solved leaves, class coverage, and semantic guards.

Two headline bridges remain: instantiating every field of the common-order criterion for each concrete V279 state, and binding the generated census to a formal decision tree on 23 items. Therefore the result is an internal exact-computation upper-bound candidate, not a fully formalized theorem or a proof that 54 is optimal.

\subsubsection{Decision-tree correctness versus path-count arithmetic}
A path count is not a decision-tree correctness proof. Every comparison branch must represent the corresponding order relation, every terminal return must have the requested rank in every total order extending that relation, and every dispatch must preserve the target rank. The global count becomes meaningful only after these semantic conditions are attached to the enumerated states.

Write a residual state as $(P,r,b)$, where $P$ is the current comparison poset, $r$ the requested zero-based rank, and $b$ the remaining comparison budget. Define $\mathcal C(P,r,b)$ recursively: a leaf is valid when its returned element has rank $r$ in every linear extension of $P$; an internal comparison of incomparable $x,y$ is valid when both child states extend $P$ by the corresponding relation and satisfy $\mathcal C$ with budget at most $b-1$. Induction on $b$ then gives
\begin{equation}
 \mathcal C(P,r,b)\Longrightarrow
 \text{every total order extending $P$ reaches a correct leaf in at most $b$ comparisons}.
\end{equation}
The base case is the universal-rank condition at a leaf. For the inductive step, every total order satisfies exactly one of $x<y$ or $y<x$, so execution enters the matching child; its budget is smaller, and the induction hypothesis supplies correctness and termination. Thus the semantic theorem depends on child-state validity and leaf ranks, not on the number of enumerated paths.

A residual problem can be represented by a finite poset and a target rank. Removing an element certified below the target decrements the target rank; removing an element certified above it leaves the rank unchanged. Canonical relabeling is valid only when it is a bijection preserving order constraints and transporting the selected item and target correctly. These simple rules explain why rank-decrement and relabel corruptions are important negative controls: either mistake can preserve graph sizes and node counts while changing the selection problem.

For a policy node with remaining budget $b$, each comparison child must have a smaller budget, and its state must extend the parent's comparison outcome. A solved leaf must satisfy the registered rank criterion. Induction on the budget then verifies termination and correctness of the residual policy. The supplied Lean stack formalizes these semantic ingredients and a large generated policy library; it does not allow the remaining concrete census-binding obligations to be omitted.

\subsubsection{Why classification compresses the difficult frontier}
The 11,208,192 repaired occurrences do not require that many separately synthesized strategies. Many residual states are equivalent after valid reduction and relabeling. Compressing them into 1,454 classes exposes a much smaller policy-design problem. The dispatcher certificate must nevertheless account for every original occurrence and verify the transport into its assigned class.

The three base-depth groups show how the resource bound composes. Depth 32 allows a 22-comparison continuation; depth 33 allows 21; depth 34 allows 20. Each composition is bounded by 54. The 15, 422, and 1,017 class counts sum to 1,454, and their occurrences sum to the repaired frontier. These equalities audit coverage and arithmetic; they do not establish the semantics of a single class transfer, which is checked separately.

Sixteen reusable policies are a further level of compression. They are not equivalent to sixteen original task instances, and their generated DAG nodes are not independent experiments. The contribution is the combination of a census, equivalence-preserving reduction, a small policy library, and a verified dispatch map. Each layer removes a different obstacle to independent replay.

\section{Six Mathematical Discovery Cases}
\label{app:mathematical-cases}
\label{app:mathematics}
This section presents the six mathematical discoveries as explicit statements with their hypotheses, constructions, lower bounds, equality conditions, and analytic completion steps. Exact finite certificates handle the large algebraic expansions, while the written arguments below supply the reusable theorem bridges. The result is a proof-oriented account of what EPOCH discovered and how each conclusion follows.

The logical spine of each argument is stated directly in its subsection so that an identity, finite census, analytic limit, or formal theorem cannot silently substitute for another. Table~\ref{tab:s-proof-gaps} gives the cross-case comparison without repeating those derivations.

\subsection{Hypercube determinant inequality and equality classification}

\begin{figure}[htbp]
\centering
\includegraphics[width=\linewidth]{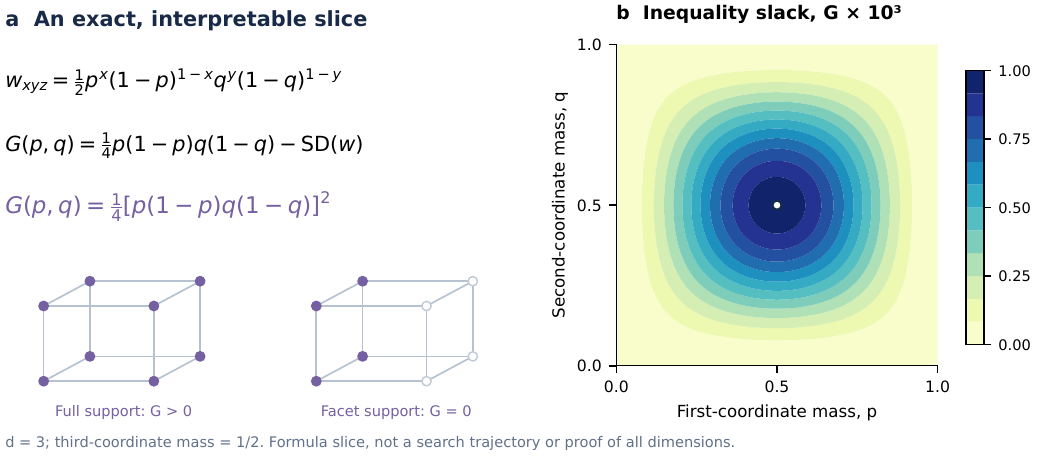}
\caption{An exact two-parameter slice of the Hypercube inequality. In dimension three take independent coordinate masses $(p,q,1/2)$, so $W=1$. The plot shows $10^3$ times the slack $G=\prod_i m_{i,0}m_{i,1}-\operatorname{SD}(w)$. Direct expansion gives $G=[p(1-p)q(1-q)]^2/4$: positive in the interior and zero on the boundary, where support lies in a facet. This formula-derived slice illustrates the all-weight, all-dimensional theorem.}
\label{fig:s-hypercube}
\end{figure}

Figure~\ref{fig:s-hypercube} provides a geometric reading of the polynomial statement and its equality condition. The support drawings illustrate the hypotheses rather than sample experimental outcomes. For this product-weight slice, an exact expansion over the $\binom84=70$ unordered vertex subsets gives the displayed slack polynomial. At $p=q=1/2$, the two sides are $15/1024$ and $16/1024$. We checked the expansion with rational arithmetic on the 25-point grid $p,q\in\{0,1/4,1/2,3/4,1\}$; this is a consistency check of the visual, not a new execution of the reported Lean proof. The general statement below does not assume product weights.

Let $w_v\geq0$ for $v\in\{0,1\}^d$, $W=\sum_vw_v$, and $m_{i,b}=\sum_{v:v_i=b}w_v$. Define the unordered determinant polynomial explicitly by
\begin{equation}
 \operatorname{SD}(w)=\sum_{\substack{U\subseteq\{0,1\}^d\\|U|=d+1}}
 \left|\det\begin{pmatrix}1&\cdots&1\\v_0&\cdots&v_d\end{pmatrix}\right|
 \prod_{v\in U}w_v,
\end{equation}
where the absolute determinant makes the ordering of $U=\{v_0,\ldots,v_d\}$ immaterial. The retained finite-polynomial theorem is
\begin{equation}
 \operatorname{SD}(w)W^{d-1}\leq\prod_{i=1}^d m_{i,0}m_{i,1}\qquad(d\geq1).
\end{equation}
For $d\geq3$, equality holds exactly when at least one facet mass is zero. For $d=1$, $W^0$ is interpreted as the polynomial constant 1, including $W=0$.

\paragraph{Proof mechanism.}
If $W>0$, normalize to a cube-valued random vector $X$. Let $\Delta$ be the affine determinant of $d+1$ independent draws. Then
\begin{equation}
 \operatorname{SD}(w)=\frac{W^{d+1}}{(d+1)!}\mathbb E|\Delta|,
 \qquad \mathbb E\Delta^2=(d+1)!\det\Cov(X).
\end{equation}
Integrality gives $|\Delta|\leq\Delta^2$, while positive semidefiniteness and Hadamard's inequality give
\begin{equation}
 \det\Cov(X)\leq\prod_i\operatorname{Var}(X_i)
 =W^{-2d}\prod_i m_{i,0}m_{i,1}.
\end{equation}
Combining the identities and restoring homogeneity proves the inequality. The $W=0$ case is immediate. Equality requires determinants in $\{0,\pm1\}$ on positive-probability tuples and diagonal covariance. The supplied formal proof excludes simultaneous equality with all facet masses positive for $d\geq3$, using rational pivots, integer clearing, and a two-level separating construction. A zero facet confines the support to a hyperplane and gives equality directly.

The reported Lean 4.19 chain proves the finite-polynomial inequality and equality characterization, including rejection of an incorrect equality mutation. The relation to the geometric conjecture of \citet{hypercubesource} still requires the complete zonoid/mixed-volume bridge and external statement-alignment review; we do not silently promote the polynomial result to every geometric formulation.

\subsubsection{Deriving the determinant second-moment identity}
Let $Z=(1,X^\top)^\top$ and let $Z_0,\ldots,Z_d$ be independent copies. Write the affine determinant as the determinant of the matrix with these columns. Expanding its square produces a double sum over permutations. Independence lets each term factor into entries of the second-moment matrix $G=\mathbb E[ZZ^\top]$. Regrouping the permutation sums yields
\begin{equation}
 \mathbb E\det(Z_0,\ldots,Z_d)^2=(d+1)!\det G.
\end{equation}
The block matrix has the form
\begin{equation}
 G=\begin{pmatrix}1&\mu^\top\\\mu&\mathbb E[XX^\top]\end{pmatrix},
 \qquad \mu=\mathbb EX.
\end{equation}
Subtracting the mean contribution through the Schur complement gives $\det G=\det(\mathbb E[XX^\top]-\mu\mu^\top)=\det\Cov(X)$. This establishes the stated identity without requiring a dimension-specific polynomial expansion.

The conversion from the ordered random sample to the unordered simplex polynomial also has a precise multiplicity. Tuples with repeated vertices contribute zero determinant. Each $(d+1)$-element vertex set appears in $(d+1)!$ orders with equal absolute determinant and product weight. Dividing by this factorial is therefore required; omitting it would alter the theorem's normalization. The powers of $W$ then restore the degree-$2d$ homogeneous inequality.

\subsubsection{Equality requires two simultaneous rigidities}
The integrality step is sharp exactly when every positive-probability determinant has absolute value zero or one. The covariance step is sharp when the relevant covariance matrix satisfies the equality condition of Hadamard's inequality. If a facet mass is zero, the support lies in a coordinate hyperplane, so all full-dimensional determinants vanish and both sides are zero.

The difficult direction is excluding equality when all facet masses are positive in dimension at least three. The first inequality constrains affine determinants of the support, while the second constrains its mixed moments. Neither condition alone excludes many nondegenerate distributions. The formal equality argument connects them through the supplied rational-pivot and separator construction. This is a separate proof component, not an automatic corollary of writing down the covariance inequality.

The low-dimensional cases are deliberately outside the stated $d\geq3$ iff classification. The inequality holds for all positive dimensions, but equality can have a different structure in dimensions one and two. Preserving the dimension qualifier prevents a true general inequality from carrying an overgeneralized equality statement.

\subsubsection{Formal statement alignment and geometric interpretation}
The Lean result checks a finite polynomial built from cube vertices, weights, and determinant coefficients. The source geometric question involves zonoids and mixed-volume formulations. Even if the finite inequality is the central algebraic obstruction, transferring the formal theorem to an external geometric claim requires the exact reduction, its normalization, and its hypotheses. A proof assistant cannot fill that bridge merely because both discussions use the word ``Hypercube.''

This case therefore has two distinct strengths: a short reusable proof mechanism for the inequality, and a reported complete formal chain for the finite statement including equality. Its remaining external gate is alignment and priority, rather than an unproved finite polynomial calculation. That status differs substantially from the kernel-only formalizations in the later analytic cases.

\subsection{H-beta front-loading: a nine-element counterexample}

\begin{figure}[htbp]
\centering
\includegraphics[width=\linewidth]{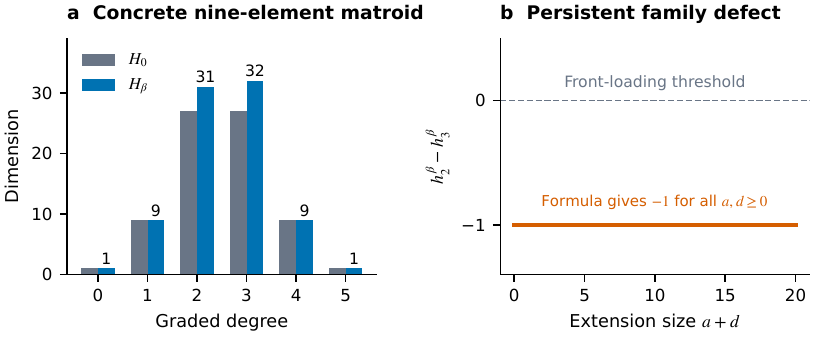}
\caption{The H-beta counterexample and its family mechanism. (a) Exact supplied graded dimensions show the middle coefficients changing from $(27,27)$ to $(31,32)$, violating the registered front-loading inequality. (b) Subtraction in the retained family formula gives $h_2^\beta-h_3^\beta=-1$ for every $a,d\geq0$; the plotted parameter range is an illustration of that identity, not an exhaustive experimental verification of the infinite family.}
\label{fig:s-hbeta}
\end{figure}

Figure~\ref{fig:s-hbeta} shows why the counterexample is structurally persistent: both middle coefficients grow by the same extension term, preserving a one-unit defect.

For the registered rank-five convention, the tested front-loading condition is $h_2^\beta\geq h_3^\beta$. The retained simple nine-element matroid $M195313$ has
\begin{align}
 \text{flat-layer counts}&=(1,9,32,49,27,1),& \#\text{bases}&=67,\\
 H_0&=(1,9,27,27,9,1),&H_\beta&=(1,9,31,32,9,1),\\
 H_\beta-H_0&=(0,0,4,5,0,0).
\end{align}
Thus $h_2^\beta-h_3^\beta=-1$, and the corresponding difference-vector defect is also $-1$. The source's differential-degree conventions must be matched explicitly when relating this computation to the conjectures of \citet{hbetasource}; the counterexample does not refute adjacent log-concavity statements.

Eight ranked cyclic flats determine all 512 subset ranks. The verification chain checks rank normalization, monotonicity, all submodularity inequalities, simplicity, flats and covers, and an independent reconstruction of the 67 bases. Exact rational atom spaces and $D_\beta$ cover matrices produce the graded dimensions through column-space recursion. Disabling $D_\beta$ changes the dimensions and is rejected. The concrete finite statement is reported in two independently structured mechanized developments.

Principal extensions along two cyclic rank-two lines produce parameters $a,d\geq0$ with retained formula
\begin{equation}
\begin{split}
H_\beta=\big(&1,9+a+d,31+6(a+d)+ad,\\
            &32+6(a+d)+ad,9+a+d,1\big).
\end{split}
\end{equation}
The middle defect remains $-1$ throughout the family. Two exact implementations support the family formula, which is not covered by the concrete Lean counterexample.

\subsubsection{From a combinatorial object to a Hilbert-series counterexample}
The counterexample begins with a matroid, not with an arbitrary proposed coefficient sequence. A sequence violating front-loading is relevant only if it is produced by the radial construction applied to a valid matroid. The verifier therefore starts with the subset-rank function and checks the matroid axioms before building the graded operators. The 512-subset domain is exhaustive for a nine-element ground set.

Rank normalization, monotonicity, and submodularity serve different purposes. Normalization fixes the empty-set rank and basic bounds; monotonicity controls inclusions; submodularity ensures the rank function has the required exchange structure. Simplicity excludes loops and parallel elements under the stated convention. The independent basis reconstruction checks that the resulting combinatorial object agrees with the reported 67 bases rather than only with a cached list of flats.

The algebraic stage builds rational atom spaces and the radial lowering operator. Its exact column-space recursion computes dimensions, which are integers but can be sensitive to a single omitted operator or an incorrect cover relation. Floating rank estimates near singularity would be inappropriate here. Exact rational arithmetic and the formal finite calculation support the displayed $H_0$, $H_\beta$, and their difference.

\subsubsection{Which conjectural property fails?}
The registered middle comparison is $31\geq32$, which is false by one. In the difference vector the analogous entries are four and five, also separated by one. These two numerical defects should not be collapsed into a claim that every proposed positivity property of the Hilbert series fails. A sequence can violate a front-loading or symmetry-related comparison while remaining log-concave.

For the concrete $H_\beta$, the neighboring log-concavity checks include $9^2>1\cdot31$, $31^2>9\cdot32$, $32^2>31\cdot9$, and $9^2>32\cdot1$. Thus the supplied coefficients themselves separate the failed middle comparison from log-concavity. The external source uses differential-degree language, so the map between that convention and the coefficient indexing in this manuscript remains an explicit alignment requirement.

\subsubsection{Why the infinite family adds information}
Write $s=a+d$. The middle entries of the extension-family formula are $31+6s+ad$ and $32+6s+ad$. Subtracting cancels every parameter-dependent contribution and leaves $-1$. The defect is therefore structurally stable under the registered principal extensions. This is stronger evidence about the failure mechanism than a single exceptional catalogue entry.

The two-parameter formula follows inductively from the extension increments. Increasing $a$ by one changes the coefficient vector by
\begin{equation}
 (0,1,6+d,6+d,1,0),
\end{equation}
while increasing $d$ by one changes it by $(0,1,6+a,6+a,1,0)$. Starting from $H_\beta(0,0)=(1,9,31,32,9,1)$, summing the first recurrence $a$ times and the second $d$ times gives
\begin{equation}
 H_\beta(a,d)=\bigl(1,9+a+d,31+6(a+d)+ad,
 32+6(a+d)+ad,9+a+d,1\bigr).
\end{equation}
Both middle coordinates receive exactly the same increment at every extension step, so their difference is the invariant $-1$. This proves persistence once the principal-extension recurrences are established from the matroid construction.

The family formula and its construction are nevertheless different evidence obligations from the concrete nine-element case. One must show that every permitted extension produces the claimed matroid structure and that the radial Hilbert computation obeys the stated formulas. The concrete Lean theorem does not quantify over these extension parameters. The manuscript attributes the family to the supplied exact implementations and written argument, while reserving full formal status for the finite counterexample.

\subsection{A nonnegative functional for reversible three-state kernels}

\begin{figure}[htbp]
\centering
\includegraphics[width=\linewidth]{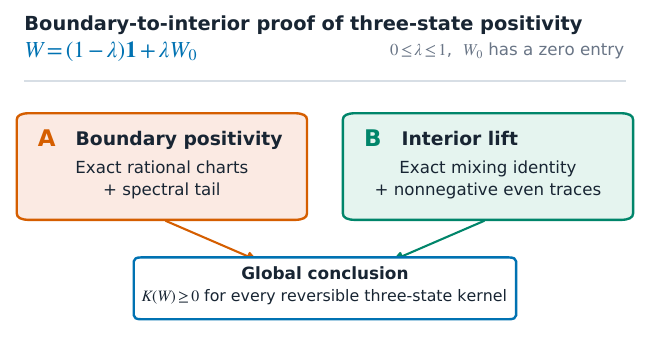}
\caption{Boundary-to-interior proof of the three-state functional. Exact rational charts and the spectral tail establish boundary positivity; the mixing identity and nonnegative even traces lift it through the interior. Together the two branches establish $K(W)\geq0$ on the registered reversible three-state domain.}
\label{fig:s-positivity}
\end{figure}

Figure~\ref{fig:s-positivity} separates the two proof interfaces. This decomposition makes it possible to inspect large exact certificates while keeping their analytic coverage obligations visible.

Let $J$ be a symmetric nonnegative $3\times3$ coupling of total mass one with positive row sums $\pi_i$. Define $P_{ij}=J_{ij}/\pi_i$ and $A=P-\mathbf1\pi^\top$. The retained theorem is
\begin{equation}
\begin{split}
 K(J)={}&2\sum_{i,j}\frac{J_{ij}}{\pi_j^2}(A^3)_{ij}(A^5)_{ij}
 +13\tr(A^8)\\
 &+16\tr(A^4)^2+252\tr(A^6)\geq0.
\end{split}
\end{equation}
For strictly positive $J$ and nonconstant normalized kernel $W_{ij}=J_{ij}/(\pi_i\pi_j)$, the retained claim is strict positivity. Related sharp-10 stability results in the source material apply to its centered rank-one and lumpable families; they are not generalized here without their full functional definitions.

\paragraph{Mixing reduction.}
Let $m=\min_{ij}W_{ij}$ and $\lambda=1-m$. For $\lambda>0$, write $W=(1-\lambda)+\lambda W_0$, where $W_0$ is nonnegative with a zero entry and the same weighted normalization. The exact identity is
\begin{equation}
\begin{split}
 K(W)={}&\lambda^9K(W_0)
 +(1-\lambda)\lambda^8\{15\tr(A_0^8)+16\tr(A_0^4)^2\}\\
 &+252(\lambda^6-\lambda^9)\tr(A_0^6).
\end{split}
\end{equation}
Here $K(W)$ denotes the coupling functional expressed through its normalized kernel. Reversibility makes the spectrum real, so the even-power traces are nonnegative. The constant-kernel endpoint is handled separately. The mechanized mixing kernel verifies this closure argument and endpoint algebra.

\paragraph{Boundary certificate.}
For the zero-diagonal boundary chart, use homogeneous parameters $(0,x,y,u,1-u,z)$ with $x=u^2(r+s)^2$, $y=(1-u)^2s^2$ after the stated chart choice. Set
\begin{equation}
 L=2u(1-u)s(r+s),\qquad U=u(1-u)((r+s)^2+s^2).
\end{equation}
For $0\leq z\leq L$, a 62,151-term polynomial certificate uses 4,478 exact pairings to absorb 3,583 negative monomials. For $L\leq z\leq U$, a 29,864-term source polynomial transforms to 141,621 positive nonzero coefficients. The $z\geq U$ spectral tail uses
\begin{equation}
 K\geq250\tr(A^6)+15\tr(A^8)+16\tr(A^4)^2\geq0.
\end{equation}
Independent implementations vary homogenization, substitution, coordinate ordering, and transforms. Corrupting pairings or a spectral coefficient is rejected. The chart discussion here is a proof outline; the supplied exact boundary package carries the full case coverage, including remaining boundary configurations. The draft does not have those complete certificate files embedded.

The result concerns three states, not arbitrary graphons or arbitrary-state rank-two kernels. Large rational boundary certificates remain outside full Lean reflection.

\subsubsection{Reversibility and spectral nonnegativity}
Let $D=\operatorname{diag}(\pi)$. Reversibility gives $DP=P^\top D$, so $D^{1/2}PD^{-1/2}$ is symmetric. The centered matrix $A=P-\mathbf1\pi^\top$ is similar to a symmetric matrix as well. Its eigenvalues are real, and for every positive integer $r$, $\tr(A^{2r})$ is the sum of nonnegative even powers. This is the spectral input used by the mixing and tail arguments.

The normalization $W_{ij}=J_{ij}/(\pi_i\pi_j)$ satisfies $\sum_j\pi_jW_{ij}=1$. Hence the minimum entry $m$ cannot exceed one. Subtracting $m$ from every entry and dividing by $1-m$ when $m<1$ gives another nonnegative normalized kernel with a zero entry. When $m=1$, weighted row normalization forces the constant kernel, which is handled as the endpoint rather than through division by zero.

Under the mixture $W=(1-\lambda)+\lambda W_0$, the centered operator scales as $A=\lambda A_0$. Most terms in $K$ therefore scale by their matrix degree. The term containing $J_{ij}$ separates into a constant-kernel contribution and a $W_0$ contribution, producing the degree-eight and degree-nine pieces in the exact formula. Tracking these degrees explains why an incorrect coefficient or exponent would invalidate the closure argument.

\subsubsection{Strictness and the limited graph-density bridge}
For a fully positive nonconstant kernel, $0<\lambda<1$ and $A_0\ne0$. Similarity to a symmetric matrix implies $\tr(A_0^6)>0$. The last mixing term is then strictly positive because $\lambda^6(1-\lambda^3)>0$. Once boundary nonnegativity is established, this yields strict interior positivity. This is a precise sufficient chain, not an appeal to numerical positivity at sampled interior points.

The supplied source relates $K$ to a particular coefficient in a graph-density expansion for $K_{5,5}-C_{10}$. Nonnegativity of that coefficient is not automatically nonnegativity of the whole graph-density difference on every kernel. The sharp-10 results use additional structure on centered rank-one or lumpable families. The manuscript preserves this boundary to avoid promoting a complete three-state coefficient theorem into general Sidorenko.

\subsubsection{What an exact positivity certificate must establish}
A rational certificate has three independent obligations. First, reconstruct the correct numerator from the original functional after clearing only denominators known to be positive in the chart. Second, prove the chart and its inequalities cover the assigned boundary region. Third, expand the proposed nonnegative blocks exactly and show that their sum equals that numerator. Positive coefficients in a polynomial produced by an incorrect substitution would not establish the theorem.

The supplied boundary proof separates zero-off-diagonal and zero-diagonal faces, then treats bounded algebraic regions and an unbounded spectral tail. The detailed sizes quoted earlier concern the zero-diagonal branch, not every face. The lower threshold $L$ and upper threshold $U$ satisfy $U-L=u(1-u)r^2\geq0$ in the registered nonnegative chart, so the middle interval is well ordered. Coverage at $u=0,1$ and other degenerate denominators must be handled by the branch contract rather than silently removed.

Weighted pairings and Bernstein transformations are certificates only after exact reconstruction. Floating optimization may find promising nonnegative block coefficients; rationalization can change identities or positivity. The independent expansion and deleted-pairing control check precisely this transition from a numerical proposal to an exact object. Full Lean reflection of these large branch certificates is still outstanding.

\subsection{Symmetric Big-centroid and Full Perazzo ranks}

\begin{figure}[htbp]
\centering
\includegraphics[width=\linewidth]{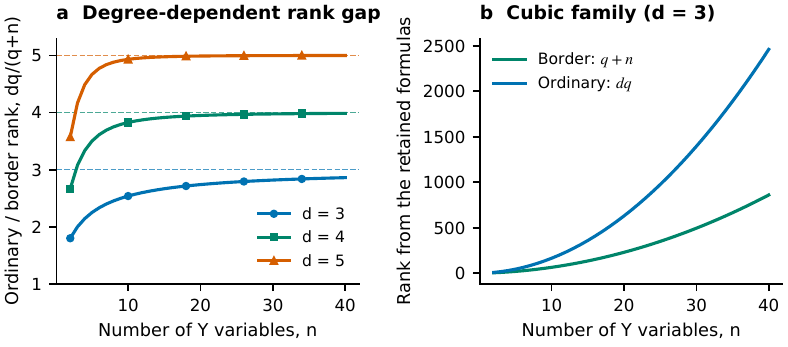}
\caption{Consequences of the retained Full Perazzo formulas over an algebraically closed characteristic-zero field, with $q=\binom{n+d-2}{d-1}$. (a) The ordinary-to-border ratio $dq/(q+n)$ approaches the degree $d$ (dashed lines) as $n$ grows. (b) For cubics, the formulas give distinct border and ordinary rank growth. Curves are calculated from the stated candidate formulas, not independently measured ranks; their validity depends on the construction and matching lower-bound arguments.}
\label{fig:s-perazzo}
\end{figure}

Figure~\ref{fig:s-perazzo} visualizes the quantitative gap that the two different rank proofs imply. The border degeneration and the ordinary-rank decomposition cannot be counted as the same object.

Over an algebraically closed field of characteristic zero, take $n\geq2$, $d\geq3$, and
\begin{equation}
 q=\binom{n+d-2}{d-1},\qquad
 B_{n,d}(X,Y)=\sum_{|\alpha|=d-1}X_\alpha Y^\alpha.
\end{equation}
The retained rank formulas are
\begin{equation}
 \brank(B_{n,d})=\brank_{\mathrm W}(B_{n,d})=q+n,
 \qquad R_{\mathrm W}(B_{n,d})=dq.
\end{equation}
Ordinary tensor border rank and symmetric Waring border rank are distinguished in the notation; the shared upper bound follows from a symmetric degeneration, with a matching flattening lower bound.

\paragraph{Cauchy-star interpolation.}
Choose pairwise distinct scalars $c_1,\ldots,c_n,t_1,\ldots,t_{n+d-2}$. Set $g(z)=\prod_i(z-c_i)$, $p(z)=\prod_j(z-t_j)$, and $A_j(Z)=\sum_i Z_i/(t_j-c_i)$. For an $(n-1)$-subset $I$ of the $t$ indices, define
\begin{equation}
 (v_I)_i=\frac{\prod_{j\in I}(c_i-t_j)}{g'(c_i)},\quad
 \ell_I(Y)=\sum_i(v_I)_iY_i,\quad
 H_I(Z)=\prod_{j\notin I}A_j(Z).
\end{equation}
Partial fractions give
\begin{equation}
 \sum_i\frac{(v_I)_i}{z-c_i}=\frac{\prod_{j\in I}(z-t_j)}{g(z)}.
\end{equation}
Hence $H_I(v_J)=0$ for $I\ne J$, while $H_I(v_I)\ne0$. The $q$ powers $\ell_I^{d-1}$ form a basis. With $f_i=p(c_i)/g'(c_i)$ and $\lambda_I=(-1)^{d-1}/H_I(v_I)$, the retained interpolation identity is
\begin{equation}
 \sum_I\lambda_I\ell_I(Y)^d=\sum_i f_iY_i^d.
\end{equation}
Solving the first-order coefficient identity for linear forms $\delta_I(X)$ yields
\begin{equation}
 B_{n,d}=\lim_{\varepsilon\to0}\varepsilon^{-1}
 \left[\sum_I\lambda_I(\ell_I+\varepsilon\delta_I)^d-\sum_i f_iY_i^d\right].
\end{equation}
The $q$ independent $X_\alpha$ derivatives and $n$ independent $Y_i$ derivatives give catalecticant rank $q+n$: the former lie in pure $Y$ degree, while the latter contain distinct $X_{(d-1)e_i}Y_i^{d-2}$ terms.

\paragraph{Ordinary Waring rank.}
The retained Schur--Kneser argument for $F=\sum_iX_iQ_i(Y)+G(Y)$ gives $R_{\mathrm W}(F)\geq d\rho$, where $\rho=\dim\operatorname{span}\{Q_i\}$. It uses vanishing second $X$ derivatives, orthogonality of coordinatewise product spaces, simultaneous stabilizer blocks, and a Kneser dimension bound in each block. For Full Perazzo, $\rho=q$. The matching upper bound decomposes each binary tangent form as
\begin{equation}
 WL^{d-1}=\sum_{j=1}^{d}\mu_j(L+t_jW)^d
\end{equation}
under the corresponding moment equations. A mechanized binary-tangent kernel proves that algebraic identity over a characteristic-zero field; it does not formalize the full Schur--Kneser lower-bound chain or guarantee every interpolation hypothesis automatically. The resulting ratio $dq/(q+n)\to d$ for fixed $d$ gives the stated obstruction to a smaller universal linear debordering coefficient on this family. External algebraic-geometric and priority review remains necessary; full border-VSP classification is not claimed.

\subsubsection{Why the interpolation family has exactly the required size}
There are $\binom{n+d-2}{n-1}=q$ subsets $I$ of size $n-1$, matching the dimension of homogeneous degree-$d-1$ forms in $n$ variables. Each $H_I$ also has degree $d-1$, because its product contains $d-1$ omitted $t$ indices. The diagonal evaluation matrix $H_I(v_J)$ therefore witnesses interpolation rather than merely supplying many candidate linear forms.

If $I\ne J$ have equal size, some index in $J$ is absent from $I$. That index occurs in the product defining $H_I$, and the partial-fraction identity makes the corresponding factor vanish at $v_J$. On the diagonal, the pairwise-distinct parameter condition prevents the necessary factors from vanishing. This explains the off-diagonal zero pattern and the nonzero diagonal without relying on a finite symbolic example.

The basis property of $\ell_I^{d-1}$ lets the $X_\alpha Y^\alpha$ target be expressed in the tangent span of the $d$th powers. The constant identity then cancels the degree-zero part. The field must allow the scalar operations and degree factors used in the tangent construction; the displayed Waring statement is therefore made in characteristic zero rather than casually transferred to small characteristic.

\subsubsection{Flattening lower bound and ordinary rank are different arguments}
The first derivative space of $B_{n,d}$ has dimension $q+n$. Derivatives with respect to the $X$ variables span all degree-$d-1$ monomials in $Y$. Derivatives with respect to the $Y$ variables have positive $X$ degree and are independent through the distinguished terms specified above. The two subspaces have different $X$ degrees and intersect trivially. A rank-one symmetric tensor contributes at most one dimension to this flattening, so the dimension gives a border-rank lower bound.

That lower bound does not produce the larger ordinary Waring rank $dq$. Ordinary rank needs the stronger product-space argument. In any proposed decomposition, vanishing second derivatives in the $X$ variables creates algebraic relations among coordinatewise products of coefficient vectors. The retained Schur--Kneser proof organizes these relations by stabilizer blocks and sums the resulting block lower bounds. Its dependencies and equality hypotheses require specialist verification; the binary tangent identity alone cannot establish the lower bound.

The upper ordinary-rank construction uses a different count: each of the $q$ tangent monomials is expressed using $d$ powers, yielding $dq$ terms. Matching this explicit upper bound with the independent lower bound yields equality. Distinguishing the two proofs prevents the border construction from being mistaken for a finite exact decomposition of length $q+n$.

\subsubsection{Binary tangent identity and debordering interpretation}
Expanding a proposed block gives
\begin{equation}
 \sum_j\mu_j(L+t_jW)^d=
 \sum_{k=0}^d\binom{d}{k}L^{d-k}W^k\sum_j\mu_jt_j^k.
\end{equation}
To obtain $WL^{d-1}$, its moments must satisfy $d\sum_j\mu_jt_j=1$ and make every other relevant moment vanish. The Lean kernel checks this implication from the supplied moment conditions. Existence of a suitable choice of weights and parameters is a separate algebraic construction, not a consequence of the implication alone.

Over an algebraically closed characteristic-zero field, that existence can be written explicitly. Let $\zeta$ be a primitive $d$th root of unity and set $t_j=\zeta^j$ and $\mu_j=\zeta^{-j}/d^2$ for $0\leq j<d$. Orthogonality of roots of unity gives
\begin{equation}
 \sum_{j=0}^{d-1}\mu_jt_j^k
 =\frac{1}{d^2}\sum_{j=0}^{d-1}\zeta^{j(k-1)}
 =\begin{cases}1/d,&k=1,\\0,&0\leq k\leq d,\ k\ne1.
 \end{cases}
\end{equation}
Substitution into the binomial expansion leaves exactly
\begin{equation}
 WL^{d-1}=\frac{1}{d^2}\sum_{j=0}^{d-1}
 \zeta^{-j}(L+\zeta^jW)^d.
\end{equation}
Thus every binary tangent block has an explicit $d$-term Waring decomposition. Applying the identity to the $q$ independent tangent blocks gives $R_{\mathrm W}(B_{n,d})\leq dq$ with no interpolation limit in this final ordinary-rank step.

For the matching lower bound, write a putative decomposition as $F=\sum_{s=1}^r L_s^d$ and split each $L_s$ into its $X$ and $Y$ parts. The vanishing of every second $X$ derivative gives, for all $X$-directions $a,b$,
\begin{equation}
 \sum_{s=1}^r a(L_s)b(L_s)L_s^{d-2}=0.
\end{equation}
The first $X$ derivatives span the $q$-dimensional space generated by the forms $Q_\alpha(Y)=Y^\alpha$. Grouping equal $Y$-directions into stabilizer blocks turns the displayed quadratic relations into orthogonality constraints on coordinatewise-product subspaces. The source Schur--Kneser lemma bounds each block by $d$ times the dimension it contributes to $\operatorname{span}\{Q_\alpha\}$; summing the blocks yields $r\geq dq$. Together with the explicit root-of-unity construction, this closes $R_{\mathrm W}(B_{n,d})=dq$ under the lemma's stated characteristic-zero hypotheses.

For fixed $d\geq3$, $q$ grows as a polynomial of degree $d-1$ in $n$, so $n/q\to0$. Consequently $dq/(q+n)\to d$. The interpretation is a family-specific obstruction to universally replacing border decompositions by ordinary Waring decompositions with a coefficient below $d$. It does not classify every minimal limiting support. The non-CB boundary counterexample in the auxiliary catalogue illustrates why that stronger geometric inference would be false.

\subsection{Heisenberg determinant limits and chain dependence}

\begin{figure}[htbp]
\centering
\includegraphics[width=\linewidth]{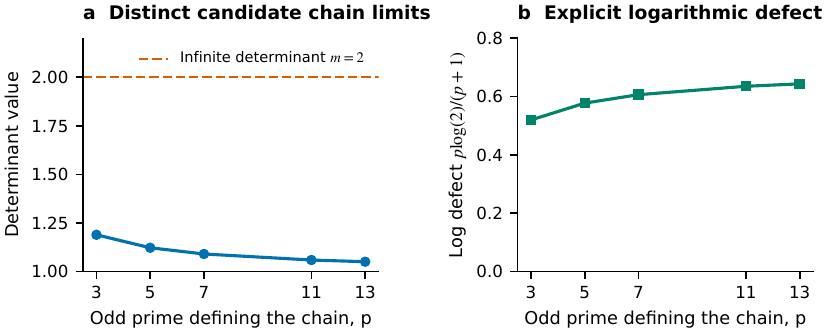}
\caption{Formula-level illustration of the Heisenberg limit candidate at $m=2$. (a) Each point is $2^{1/(p+1)}$ for a different odd-prime principal chain; the dashed line is the infinite Fuglede--Kadison determinant, 2. (b) The corresponding log defect is $p\log(2)/(p+1)$. Lines connect selected prime parameters only as visual guides. These are candidate limiting values, not finite-$q$ measurements or a numerical convergence test.}
\label{fig:s-heisenberg}
\end{figure}

Figure~\ref{fig:s-heisenberg} exposes the claimed chain dependence without extrapolating the small finite determinant controls. Establishing each limit still requires uniform packet estimates.

In the integer Heisenberg group, set $A_m=1-ma+mb$ for integer $m\geq2$. For finite principal quotients at $q=p^i$, define the reported normalization by $D_q(A_m)=|\det A_{m,q}|^{1/q^3}$ when the finite determinant is nonzero. Under the retained principal-chain hypotheses with odd prime $p$, the candidate limit is
\begin{equation}
 D_q(A_m)\longrightarrow m^{1/(p+1)},
\end{equation}
whereas the cited infinite Fuglede--Kadison determinant is $m$. The logarithmic defect is $p\log m/(p+1)$. The designated $(p,m)=(3,2)$ special case gives $2^{1/4}$. This is presented as an exact-limit follow-up to the determinant-approximation counterexample of \citet{heisenbergsource}, not as rediscovery of that original counterexample.

\paragraph{Finite representation packets.}
For exact central order $d\mid q$, put $n=q/d$. Irreducible packets have dimension $d$ and parameters $(u,v)\in\mu_n^2$, with regular-representation dimension check
\begin{equation}
 \sum_{d\mid q}\varphi(d)n^2d^2=q^3.
\end{equation}
For odd-order packets, the quantum-binomial and trace identities for $T=\rho(a)-\rho(b)$ give $T^d=(u-v)I$ and the reverse characteristic polynomial, yielding
\begin{equation}
 \det(I-mT)=1-m^d(u-v).
\end{equation}
The mechanized finite-packet kernel proves the implication from the reverse characteristic polynomial to this determinant identity. Representation classification, multiplicities, packet hypotheses, and the complete resultant product remain external to that kernel.

\paragraph{Limit mechanism and controls.}
Complete Galois packets combine all central orders. Small $d$ use a nonzero integer norm for a lower bound; large $d$ use main-term dominance and root-of-unity spacing. Uniform normalized error estimates then supply the asserted limit. These analytic estimates are a necessary proof layer; finite determinant agreement alone does not establish convergence.

Direct integer matrices and resultant controls include $(q,m)=(3,2),(3,3),(5,2),(9,2),(2,2),(4,2),(6,2),(8,2)$, with largest matrix dimension 729. The determinant is 1,729 at $(3,2)$ and $-735$ at the even control $(2,2)$, requiring an absolute-determinant convention. Even controls use their appropriate packet rules, not an unchecked extension of the odd-order identity. Follow-up chain classifications, nonconvergence constructions, and cofinal recovery claims remain within this specialist research line; the current manuscript does not treat them as additional independent discoveries.

\subsubsection{Regular-representation multiplicities and determinant products}
The finite quotient has order $q^3$. Its regular representation contains each irreducible representation with multiplicity equal to its dimension. Therefore an irreducible packet determinant must be raised to that multiplicity when reconstructing the full determinant. The identity $\sum_{d\mid q}\varphi(d)n^2d^2=q^3$ checks the dimension accounting, but a complete reconstruction must also ensure that the packets are neither omitted nor duplicated.

Writing $\mathcal P_d$ for the packets of exact central order $d$ and $\Delta_{d,u,v}=1-m^d(u-v)$, the regular determinant has the product form
\begin{equation}
 |\det A_{m,q}|=
 \prod_{d\mid q}\prod_{(u,v)\in\mathcal P_d}|\Delta_{d,u,v}|^d.
\end{equation}
Consequently its normalized logarithm is
\begin{equation}
 \log D_q(A_m)=\frac{1}{q^3}
 \sum_{d\mid q}d\sum_{(u,v)\in\mathcal P_d}
 \log|1-m^d(u-v)|.
\end{equation}
This formula identifies exactly what the limit proof must control: packet multiplicities, the main $d\log m$ contribution, and the logarithmic error caused by possible proximity of $u-v$ to $m^{-d}$.

The scalar identity $\det(I-mT)=1-m^d(u-v)$ simplifies each packet dramatically. It does not itself determine how frequently a given $(d,u,v)$ occurs. This is why packet algebra and representation classification are listed as separate verification layers. Incorrect multiplicities can preserve each packet identity while giving the wrong normalized finite determinant and limiting exponent.

The reverse characteristic polynomial is convenient because it encodes $\det(I-zT)$ directly. If it equals $1-cz^d$, substituting $z=m$ proves the packet formula. The Lean kernel formalizes this algebraic implication. The quantum-binomial relation and vanishing lower traces are the mathematical route to its hypothesis in the registered representation family; they remain to be connected formally for the full theorem.

\subsubsection{Why exact finite controls do not establish the limit}
Agreement between direct integer determinants and resultants at $q\leq9$ tests packet reconstruction, signs, and normalization in concrete cases. It cannot prove convergence as $i\to\infty$. A limiting formula needs an error estimate that remains valid over all packets and all levels in the chain. Near-cancellation in factors of the form $1-m^d(u-v)$ is especially relevant after taking logarithms.

The retained proof divides the packets into small and large central orders. Complete Galois products at small orders are nonzero integers, giving a lower bound that would not be available for an isolated small complex factor. At large orders, dominant terms and spacing among roots of unity control the normalized error. The cutoff and uniform estimates are part of the analytic proof, not parameters chosen to fit the finite determinant values.

More explicitly, choose a cutoff $D=D(q)$ with $D\to\infty$ and $D=o(q)$. For $d\leq D$, combine every Galois orbit before taking logarithms. Nonvanishing of the resulting integer norm prevents an individual near-zero complex factor from producing an uncontrolled negative contribution; the total small-order contribution is $o(q^3)$ after normalization. For $d>D$, factor
\begin{equation}
 1-m^d(u-v)=m^d(u-v)
 \left(1-\frac{1}{m^d(u-v)}\right).
\end{equation}
The first factor supplies $d\log m$ plus the root-of-unity term $\log|u-v|$. The spacing estimate controls the latter after complete packet summation, while $m^{-d}$ makes the last logarithm uniformly summable away from the paired zero modes. Under the retained packet-counting lemma, the coefficient of $\log m$ converges to $1/(p+1)$ and every residual sum is $o(1)$ after division by $q^3$. Exponentiating therefore gives $D_q(A_m)\to m^{1/(p+1)}$.

Taking logarithms of the claimed limit gives $(\log m)/(p+1)$, while the infinite determinant has logarithm $\log m$. Their difference is $p\log m/(p+1)$. This elementary consequence makes the size of the defect explicit but does not replace the convergence argument. It also emphasizes chain dependence: a different quotient chain may have a different packet distribution and a different limit or no limit.

\subsubsection{Sign conventions and specialist review}
The negative even-level determinant $-735$ is a useful convention check. A normalized determinant intended to compare with a positive Fuglede--Kadison determinant must use the registered absolute or appropriate singular-value convention. Copying the odd-order packet formula into an even-level test without adjusting its hypotheses would be a separate error. The manuscript reports the even controls precisely to make such convention failures visible.

The source counterexample demonstrates the failure of determinant approximation. The exact limit and chain classifications are subsequent results that call for an independent operator-algebraic analysis. Although the formal determinant identity provides substantial and reusable evidence, the principal analytical challenge lies beyond the identity itself.

\subsection{Online permutation embedding: alternating-family asymptotics}

\begin{figure}[htbp]
\centering
\includegraphics[width=\linewidth]{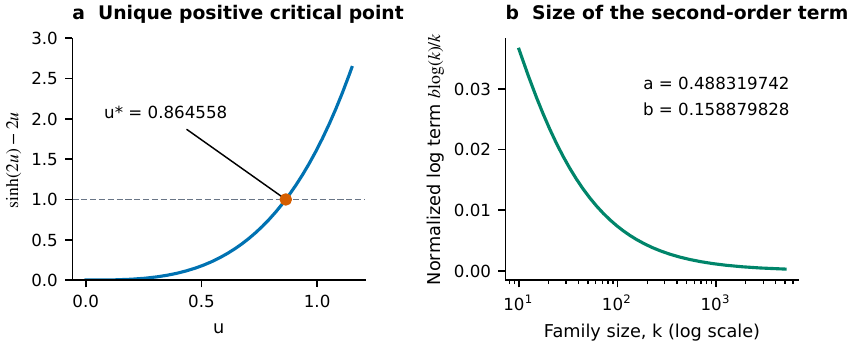}
\caption{Analytic structure of the alternating-family expansion. (a) The critical equation $\sinh(2u_*)-2u_*=1$ determines $a=\cosh^2(u_*)/4$ and the logarithmic coefficient. (b) The plotted quantity is only the explicit correction $b\log(k)/k$ to $\beta(\operatorname{Alt}_k)/k^2$, where $b=\sinh^2(u_*)/6$; the unknown $O(k)$ remainder in the unnormalized expansion is not plotted. These are formula illustrations, separate from the reported Arb dynamic-programming checks.}
\label{fig:s-online}
\end{figure}

Figure~\ref{fig:s-online} connects the scalar critical point to the scale of the logarithmic term. It does not plot new dynamic-programming observations or assert global optimality of the alternating family.

Let $\operatorname{Alt}_k$ denote the alternating increasing family and $\beta$ the objective of the source online embedding problem \citep{onlinesource}. Let $u_*>0$ solve $\sinh(2u_*)-2u_*=1$. The retained asymptotic candidate is
\begin{equation}
 \frac{\beta(\operatorname{Alt}_k)}{k^2}\longrightarrow
 \frac{\cosh^2u_*}{4}\approx0.49.
\end{equation}
More precisely, for parity constants $C_0,C_1$,
\begin{equation}
\begin{split}
 \sqrt{\beta(\operatorname{Alt}_k)}={}&\frac{\cosh u_*}{2}k
 +\frac{\sinh^2u_*}{6\cosh u_*}\log k+C_{k\bmod2}+o(1),\\
 \beta(\operatorname{Alt}_k)={}&\frac{\cosh^2u_*}{4}k^2
 +\frac{\sinh^2u_*}{6}k\log k+O(k).
\end{split}
\end{equation}

\paragraph{Recurrence mechanism.}
The alternating-tree root splits into one singleton and an alternating subtree of size $k-2$, giving $S_k=\Phi(S_{k-2})$ for $S_k=\sqrt{\beta(\operatorname{Alt}_k)}$. The retained Bellman-residual expansion is
\begin{equation}
 P_{1,s^2}(u)=\sinh(2u)-2u-\frac{8\sinh^3u}{3s}+O(s^{-2}).
\end{equation}
The root condition yields $u_s=u_*+2\sinh(u_*)/(3s)+O(s^{-2})$, and hence
\begin{equation}
 \Phi(s)-s=\cosh u_*+\frac{\sinh^2u_*}{6s}+O(s^{-2}).
\end{equation}
Summation along each parity class gives the linear and logarithmic terms. This step requires uniform error control in the Bellman expansion, beyond the formal algebraic kernel.

With $x=e^{-2u}$, the variational objective becomes $H(u)=(1+2u)\coth(u)/8$. Its derivative has the sign of $\sinh(2u)-2u-1$, giving a unique minimizer $u_*$ and $H(u_*)=\cosh^2u_*/4$. A mechanized scalar kernel proves the critical-equation rewrite and coefficient normalization. Root uniqueness and recurrence asymptotics are not included in that scope.

Two outward-rounded Arb dynamic programs support finite optimality of the alternating family within the frozen finite problem for $k\leq5000$. The reported finite second-order estimates both round to 0.11 and approach the exact expression $\sinh^2(u_*)/(6\cosh(u_*))\approx0.11$, the logarithmic coefficient in $\sqrt{\beta}$; the $k\log k$ coefficient in $\beta$ is instead $\sinh^2(u_*)/6\approx0.16$. This agreement links the finite dynamic programs to the analytic expansion.

\subsubsection{The critical equation and its unique positive solution}
Define $f(u)=\sinh(2u)-2u-1$ for $u\geq0$. Then $f(0)=-1$, $f(u)\to\infty$, and $f'(u)=2\cosh(2u)-2>0$ for $u>0$. Thus continuity and strict monotonicity give exactly one positive root. This elementary argument is a written derivation here; the supplied Lean kernel is reported only for the critical identity and coefficient normalization.

For $H(u)=(1+2u)\coth(u)/8$, differentiation gives
\begin{equation}
 H'(u)=\frac{\sinh(2u)-2u-1}{8\sinh^2u}.
\end{equation}
The denominator is positive for $u>0$, so $H$ decreases before $u_*$ and increases after it. At the root, $1+2u_*=2\sinh u_*\cosh u_*$, and substitution yields $H(u_*)=\cosh^2u_*/4$. This connects the scalar equation, the variational minimizer, and the leading scaling coefficient.

\subsubsection{Extracting the logarithmic correction from the recurrence}
Let $a=\cosh u_*$ and $b=\sinh^2u_*/6$. The registered expansion has the form $S_k-S_{k-2}=a+b/S_{k-2}+O(S_{k-2}^{-2})$. On either parity class, set $x_j=S_{2j+\eta}$. Then
\begin{equation}
 x_{j+1}-x_j=a+\frac{b}{x_j}+O(x_j^{-2}).
\end{equation}
Once the recurrence establishes $x_j\sim aj$, the reciprocal term contributes $(b/a)\log j$ on summation. The quadratic error is summable. Controlling the difference between $1/x_j$ and $1/(aj)$ refines the result to $x_j=aj+(b/a)\log j+C_\eta+o(1)$ under the uniform expansion hypotheses.

Here is the summation step in detail. The recurrence first gives constants $c_1,c_2>0$ with $c_1j\leq x_j\leq c_2j$ for all large $j$, hence $x_j=aj+O(\log j)$ by summing $b/x_j+O(x_j^{-2})$. Therefore
\begin{equation}
 \frac1{x_j}=\frac1{aj}+O\!\left(\frac{\log j}{j^2}\right).
\end{equation}
Define $y_j=x_j-aj-(b/a)\log j$. Using
\begin{equation}
 \log(j+1)-\log j=\frac1j+O(j^{-2}),
\end{equation}
the recurrence gives $y_{j+1}-y_j=O((\log j)/j^2)$. The series of these increments converges absolutely, so $y_j$ converges to a parity-dependent constant $C_\eta$. This proves
\begin{equation}
 x_j=aj+\frac ba\log j+C_\eta+o(1)
\end{equation}
from the uniform one-step expansion, rather than inferring the logarithmic term from numerical fitting.

Returning to $k$ gives $S_k=(a/2)k+(b/a)\log k+C_{k\bmod2}+o(1)$, with the parity-dependent constants absorbing fixed logarithmic shifts. Squaring produces the cross term $b\,k\log k$, while the squared logarithm and constant interactions are $O(k)$. This recovers exactly the coefficient $\sinh^2u_*/6$ in the displayed expansion for $\beta$.

The derivation clarifies which parts are formal algebra and which parts require asymptotic control. The critical-value simplification is a finite identity. Establishing the Bellman residual expansion uniformly, selecting the appropriate root branch, proving $S_k\to\infty$, and summing errors are analytic obligations. They cannot be discharged merely by checking the coefficient identity in Lean.

\subsubsection{Finite optimality, family asymptotics, and global extremality}
The Arb calculations address the frozen finite optimization problem through horizon 5,000, using outward rounding so that numerical comparisons are certified as intervals. The two implementations reduce the chance of a shared dynamic-programming error. Their agreement on a finite range does not show that the alternating family is globally optimal at every larger size.

The asymptotic statement concerns the alternating family itself. Even a complete proof of that family recurrence and its limit would not establish that no other family has a smaller limiting objective. The global extremal question requires a comparison principle over all allowed permutations or trees. The manuscript keeps these three levels distinct: finite certified comparisons, the family's analytic expansion, and the unresolved all-family extremality claim.

The second-order finite estimates approaching $\sinh^2(u_*)/(6\cosh(u_*))\approx0.11$ are consistency checks at the $\sqrt{\beta}$ scale. The separate coefficient $\sinh^2(u_*)/6\approx0.16$ belongs to the $\beta$ scale. Keeping the exact symbolic expressions alongside two-decimal approximations prevents those two coefficients from being conflated.

\clearpage
\section{Auxiliary Results: Exact Classifications, Reductions, and Obstructions}
\label{app:auxiliary-results}
\label{app:catalogue}
The thirteen auxiliary groups in this section comprise twenty-five scoped results that expand the supporting research inventory. They are not counted as additional independent 4+6 headline successes. They include finite classifications, infinite-family candidates, verification improvements, source-table corrections, and negative local obstructions. The statements are reproduced and organized from the supplied research dossier. When source-specific notation is not fully defined in that dossier, the statement retains its source convention and identifies the additional definition needed for standalone specialist assessment.

\subsection{Exact k-SSD heights and unique extremizers}
For a positive-integer $k$-SSD set $A=\{a_1,\ldots,a_n\}$, all sums $\sum_i c_ia_i$ with $c_i\in\{0,\ldots,k\}$ are distinct. In the supplied convention, the height is the largest generator, and $h_k(n)$ minimizes that height. A candidate upper bound is easy to certify by enumerating coefficient sums. An exact minimum requires excluding every smaller admissible generator tuple; uniqueness additionally requires showing that no other tuple attains the minimum.
\begin{table}[htbp]
\centering\small
\caption{Dual-verified four-generator k-SSD results.}
\begin{tabular}{@{}lrlr@{}}
\toprule
Parameter & Height & Unique extremizer & Coefficient sums \\
\midrule
$h_4(4)$ & 94 & $\{73,89,92,94\}$ & 625 \\
$h_5(4)$ & 168 & $\{135,156,166,168\}$ & 1,296 \\
$h_6(4)$ & 254 & $\{214,247,251,254\}$ & 2,401 \\
\bottomrule
\end{tabular}
\end{table}
The two verification routes use signed-difference frontiers and translate-disjointness enumeration. Their agreement covers lower slices and extremizer uniqueness, not merely validity of the displayed sets. The supplied comparator construction gives heights 121, 211, and 337, respectively. These are scoped finite improvements, not an all-$k$ height formula.

The primary search also reports $h_4(5)=442$ with unique extremizer $\{344,417,427,433,442\}$ after 37,759,246,033 fifth-value tests. Its independent Z3 lower exclusion timed out. This result must remain primary-complete rather than dual-verified. The timeout does not falsify the candidate; it prevents the stronger independent-exclusion status. This entry illustrates why construction, lower-bound exclusion, and uniqueness need separately recorded receipts.

\subsection{Four-element sumsets of diameter six and seven}
Let $hA$ denote the set of sums of exactly $h$ elements of $A$, with repetition, and define $R(h,4)=\{|hA|:A\subset\mathbb Z,\ |A|=4\}$. The retained diameter-six classification states that for $h\geq4$,
\begin{equation}
R(h,4)\cap[5h+2,7h-6]=
\begin{cases}
\varnothing,&h=4,\\
\{6h-3,6h-1\},&h=5,6,\\
\{6h-3,6h-1,6h+1\},&h\geq7.
\end{cases}
\end{equation}
The proof reduces to affine-normalized diameter six and verifies exact complement identities for five reflection representatives. Affine normalization removes translation and common scaling, while reflection identifies equivalent endpoint patterns.

The diameter-seven region has, for $h\geq6$,
\begin{equation}
R(h,4)\cap[6h+2,8h-9]=
\begin{cases}
\{7h-4,7h-3\},&h=6,\\
\{7h-5,7h-4,7h-3,7h-2\},&7\leq h\leq9,\\
\{7h-5,7h-4,7h-3,7h-2,7h+1\},&h\geq10.
\end{cases}
\end{equation}
Nine normalized reflection representatives cover this second reduction. Independent Presburger queries verify the arbitrary-$h$ induction steps. Finite checks through 200 are controls, not the reason the statements quantify over every larger $h$. The important evidence chain is structural diameter reduction, a finite list of normalized forms, and symbolic induction on each form. External priority for these exact classification windows remains separate.

\subsection{Rajagopal finite boundary through h+k at most thirteen}
Using the source convention $R(h,k)=\{|hA|:A\subset\mathbb Z,\ |A|=k\}$, the retained constructive verification covers all registered pairs $k>h$ with $h+k\leq13$. The new boundary is
\begin{align}
 R(4,9)&=[33,495]\setminus\{34,35,38\},\\
 R(5,8)&=[36,792]\setminus\{37,38,39,42,43,47\},\\
 R(6,7)&=[37,924]\setminus\{38,39,40,41,44,45,46,50,51,56\}.
\end{align}
The boundary search emits 460, 751, and 878 witnesses. Their sum is 2,089; this is the new boundary-layer count, distinct from the broader inventory's 2,814 witnesses across its registered range. An independent program reconstructs the sumsets of all 2,089 boundary witnesses and rejects duplicate-element corruptions.

Existence witnesses establish every included value. Exact interval closure and the source's exclusion argument are separately needed to establish that omitted values are impossible. The dossier reports the full finite verification, but the raw witness catalogue and exclusion certificates are not embedded in this manuscript. The claim is a finite boundary theorem, not a proof of the all-parameter conjecture. This entry demonstrates the value of retaining constructive witnesses instead of a table of possible cardinalities alone.

\subsection{Lonely Runner single-replacement classifications}
For the source's single-replacement speed families with deleted speed $r=4$ or $r=5$, the retained classification finds no sporadic tight cases outside the published multiple-replacement mechanism. In the $r=5$ reduction, exactly 46 candidate pairs remain with $11\leq n\leq20$. Their exact Breakpoint-Lemma evaluations satisfy $\operatorname{LR}>1/n$.

The smallest retained margin is $1/42$, attained at the source parameters $(n,w)=(14,14)$ with $\operatorname{LR}=2/21$; indeed $2/21-1/14=1/42$. This arithmetic is an internal consistency check on the stated margin. The source definition of the replacement family and the proof that these 46 pairs exhaust its non-multiple cases remain part of the theorem's hypotheses and reduction package.

Combining the finite residual classification with the established multiple case gives the stated fixed-$r=5$ result. The finite computations alone would not imply it without that reduction. Neither classification resolves the general Lonely Runner conjecture. The separate unresolved 14-runner solver direction appears in the negative ledger and must not be confused with this admitted fixed-replacement family.

\subsection{Fourier principal nonsingularity for orders 10p and 11p}
The source lifting theorem reduces two prime-parameter Fourier nonsingularity families to exact cyclotomic norm data. For order $11p$, the retained exception support is $\{3,11,23,67,199,419\}$ and the aggregate norm factorization is
\begin{equation}
 3^{220}11^{28160}23^{1100}67^{220}199^{220}419^{220}.
\end{equation}
Thus the stated principal-minor nonsingularity conclusion holds for primes outside that support under the source theorem's admissibility hypotheses; order 187 is included. For order $10p$, the support is $\{2,3,5,11,31\}$ with exponents $(5120,120,5120,240,80)$, and order 130 is included.

Multiplication-matrix determinants computed by Bareiss elimination and independent resultants agree over all 2,048 order-11 subsets and 1,024 order-10 subsets in the retained enumeration. Their independent representations help detect normalization or cyclotomic arithmetic errors. An exceptional prime in the norm support means the sufficient lifting conclusion is unavailable there; it should not automatically be described as a proved singular case.

\subsection{ERG(3,8): cycle spectrum and factor-criticality}
Conditional on completeness and correctness of the frozen 477,142-record catalogue, every listed triangle-free 27-vertex graph with independence number at most seven has cycles of every length from 4 through 27 and is factor-critical. Here factor-criticality means that deleting any vertex leaves a graph with a perfect matching, consisting of 13 edges in this case.

The primary pass constructs cycles and vertex-deleted matchings. The independent pass uses generic cycle enumeration and a skew Tutte matrix over $\mathbb F_{2147483647}$. Rank 26 with a full-support kernel gives the required nonvanishing principal cofactors in the odd-order skew setting, which certifies the vertex-deleted matching condition under the checker theorem. The finite-field certificate is a witness of nonvanishing, not an inference from a failed search for a matching.

\subsection{Kasami APN counting identity at n equal to fourteen}
For admissible $k$ with $\gcd(k,14)=1$, the retained finite result verifies the source NSUCRYPTO counting identity for the Kasami power $F(x)=x^{4^k-2^k+1}$ over $\mathrm{GF}(2^{14})$. Frobenius symmetry reduces the new representative computations to $k=3$ and $k=5$.

Two different irreducible-polynomial field models check 16,382 admissible $\rho$ values per representative, giving 32,764 exact identities $Z(\rho)=0$ in the source notation. Both derivative images have 8,192 elements with two-point fibers, and the reported Walsh support has size 3,088. 

The formula defining $Z$ and the symmetry reduction are part of the source problem and must be included with a standalone certificate release. The present evidence extends a finite verified boundary from $n\leq13$ to $n=14$. It does not establish an arbitrary-$n$ counting theorem or a new general APN construction. The work illustrates exact finite algebra at a new parameter boundary with representation-independent checks.

\subsection{ReLU MAX8 certificate compression and subset-profile dynamic programming}
Starting from the public four-template MAX6 certificate, the local attachment search contains 676 raw candidates, 127 graph-isomorphism classes, and 125 exact function columns. The retained combination yields a 25-template degree-three MAX8 identity. The source's public 69-template certificate expands to 515,488 symmetric second-layer atoms, whereas the retained certificate uses 187,460: a reported 63.63\% reduction and 2.75-fold compression.

Full $8!$ permutation enumeration and subset-profile dynamic programming agree term by term. The dynamic program aggregates permutations according to profiles sufficient for the certificate evaluation, so its correctness requires proving that aggregated transitions preserve every evaluated contribution. Agreement on a complete finite instance checks that reduction against brute-force enumeration.

The same generic verifier checks public MAX$_N$ certificates for $5\leq N\leq10$, reducing the largest replay from approximately 1.46 billion permutations to 71.77 million transitions. This is a verification-efficiency result as well as a compressed certificate.

\subsection{AES CNOT Pareto cleanup}
A separate local rewrite begins with the source 20-qubit AES circuit containing 44 Toffoli gates and 197 CNOT gates. Two identical CNOT pairs can commute through intervening gates and cancel, giving 193 CNOT gates while preserving all 44 Toffolis. The two retained commutation windows span 22 and 23 gates, and the independent verifier checks all 47 intervening commutations recorded by the certificate.

All 256 input cases pass the independent truth-table check. This validates the finite circuit transformation under the registered input and ancilla contract. It is a modest Pareto improvement: one gate type decreases while the dominant Toffoli count stays fixed. It is distinct from the Hyperfactor cross-output construction and should not be added to its T-count savings as though both changes were applied to the same baseline.

Separate exhaustive single-delete and 10,920 two-delete/one-insert affine charts do not reduce the eight-Toffoli count of the four-bit inverse. Those search negatives prove only absence within the explored rewrite neighborhoods. They are not a global Toffoli lower bound.

\subsection{Factorwise tail elimination for transferred tensors}

For st333, EPOCH identified a retained optimizer in which one derivative mode vanishes in each of ten rank-one terms, eliminating the cubic derivative product. An exhaustive search over all $3^{10}=59{,}049$ mode patterns found 1,536 admissible realizations; the selected path has coefficient weights $(0,18,8,0)$ and gives
\begin{equation}
 R(T^{\otimes N})\leq (N+1)10^N.
\end{equation}
Under the supplied scalar-count model, this lowers the crossover with exact rank-11 recursion from $N=48$ to $N=39$, and the crossover with explicit base-field recursion from $N=90$ to $N=81$.

The same mechanism transfers to ss333 and to the 35-term st555 construction. For st555, the coefficient weights become $(0,75,922,0)$, improving the two reported crossover thresholds from $(41,78)$ to $(33,70)$. This session shows how EPOCH can convert a structural observation about a border construction into a measurable reduction in exactification overhead.

\subsection{Degree-eight Smale reductions}

EPOCH reduced the nearest-critical-point problem to a one-variable polynomial problem on the unit disk using the Grace--Walsh--Szeg\H{o} theorem. Exact Schur and Sturm certificates then establish
\begin{equation}
 \left|\frac{P(b)}{b}\right|>\frac{433}{5000}\approx 0.09,
\end{equation}
which improves the corresponding general-degree estimate at degree eight by a factor of approximately $4.31$ under the same normalization.

The analysis also isolates the multiple-critical-point case and proves the sharp bound
\begin{equation}
 \left|\frac{P(b)}{b}\right|\geq \frac18,
\end{equation}
with equality attained by $[1-(1-z)^8]/8$. The session combines a classical reduction with exact region certification, yielding both a stronger numerical bound and a sharper structural description of any remaining counterexample.

\subsection{A cofinite unfair-trinomial family}

EPOCH turned a finite recurrence pattern into a parameter-uniform obstruction. For every integer $k\geq 3{,}000{,}001$ with $\gcd(k,3)=1$ and every $0<a<1$, no nonzero polynomial with nonnegative coefficients can multiply
\begin{equation}
 1+ax^3+x^k
\end{equation}
into a polynomial whose coefficients all belong to $\{0,1\}$.

The proof organizes the recurrence into three residue classes and controls the resulting negative-binomial packets through projective contraction, seam estimates, an antiperiodic theta kernel, and a terminal recurrence. Exact symbolic and rational replays verify the algebraic identities used throughout. This case illustrates EPOCH's capacity to recognize a computational pattern, identify the mechanism behind it, and promote it to an explicit theorem for an infinite family.

\subsection{Sharp stability and an exact loopless mixing curve}

For the loopless three-state reversible base $W_0$, let $W_t=(1-t)+tW_0$ for $0\leq t\leq1$, and let $H=K_{5,5}-C_{10}$. EPOCH obtained the exact inequality
\begin{equation}
 t(H,W_t)\geq c(t)\,t(C_4,W_t)^5,
 \qquad
 c(t)=\frac{2h(t)}{(8+t^4)^5},
\end{equation}
where
\begin{align}
 h(t)={}&16384+10240t^4+7680t^6-1280t^7+5760t^8-3200t^9\nonumber\\
       &+2656t^{10}-1440t^{11}+440t^{12}-60t^{13}-t^{15}.
\end{align}
The ratio increases from $1$ to $34/27$, with equality for $t>0$ precisely at the uniform triangle, up to the stated scaling and relabelling convention.

Global monotonicity is certified by 6,615 positive ordered Bernstein controls, 13,140 positive ratio controls, and a reduced derivative certificate with 14 positive controls. A complementary sharp-stability argument is represented by 31,765 rational nonnegative blocks. This session demonstrates EPOCH's ability to organize a large exact certificate into a transparent quantitative theorem with a sharp equality case.

Across these sessions, the advantage of EPOCH is not the volume of computations performed, but the conversion of search output into reusable mathematical structure: optimized constructions, exact inequalities, uniform parameter arguments, and independently replayable certificates.

\ifdefined\EPOCHStandaloneSupplement
\bibliography{references}

\begin{thebibliography}{28}
\providecommand{\natexlab}[1]{#1}
\providecommand{\url}[1]{\texttt{#1}}
\expandafter\ifx\csname urlstyle\endcsname\relax
  \providecommand{\doi}[1]{doi: #1}\else
  \providecommand{\doi}{doi: \begingroup \urlstyle{rm}\Url}\fi

\bibitem[{AlgoTune contributors}(2026)]{algotune}
{AlgoTune contributors}.
\newblock {AlgoTune}: Official task repository, 2026.
\newblock URL \url{https://github.com/oripress/AlgoTune}.
\newblock Pinned task and data revisions are reported in the supplementary
  material; accessed September 17, 2026.

\bibitem[Altschuler et~al.(2026)Altschuler, Dubroff, and
  Tikhomirov]{onlinesource}
Dylan~J. Altschuler, Quentin Dubroff, and Konstantin Tikhomirov.
\newblock Online permutation embedding: Optimal stopping and scaling laws.
\newblock \emph{arXiv preprint arXiv:2608.19050}, 2026.
\newblock URL \url{https://arxiv.org/abs/2608.19050}.

\bibitem[{Anthropic}(2026)]{clauderiemann}
{Anthropic}.
\newblock Learning more about {Claude}'s mathematical capabilities, August
  2026.
\newblock URL \url{https://www.anthropic.com/research/riemann-zeta}.
\newblock Primary research announcement, August 10; revised August 13, 2026.
  Reports a lower bound for the proportion of critical-line zeros, not a proof
  of the Riemann hypothesis. Accessed September 18, 2026.

\bibitem[Averkov et~al.(2026)Averkov, von Dichter, and
  Soprunov]{hypercubesource}
Gennadiy Averkov, Katherina von Dichter, and Ivan Soprunov.
\newblock On the log-submodularity for zonoids: from mixed volume inequalities
  to the hypercube.
\newblock \emph{arXiv preprint arXiv:2608.14909}, 2026.
\newblock URL \url{https://arxiv.org/abs/2608.14909}.

\bibitem[Cemri et~al.(2026)Cemri, Agrawal, Gupta, Liu, Cheng, Mang, Naren,
  Erdogan, Sen, Zaharia, Dimakis, and Stoica]{adaevolve}
Mert Cemri, Shubham Agrawal, Akshat Gupta, Shu Liu, Audrey Cheng, Qiuyang Mang,
  Ashwin Naren, Lutfi~Eren Erdogan, Koushik Sen, Matei Zaharia, Alex Dimakis,
  and Ion Stoica.
\newblock {AdaEvolve}: Adaptive {LLM} driven zeroth-order optimization.
\newblock \emph{arXiv preprint arXiv:2602.20133}, 2026.
\newblock URL \url{https://arxiv.org/abs/2602.20133}.

\bibitem[D{\"o}rrer et~al.(2025)D{\"o}rrer, Gendle, Betz, von Smercek, Steding,
  and Stober]{selection}
Josua D{\"o}rrer, Konrad Gendle, Johanna Betz, Julius von Smercek, Andreas
  Steding, and Florian Stober.
\newblock Exact lower bounds for the number of comparisons in selection.
\newblock In \emph{23rd International Symposium on Experimental Algorithms (SEA
  2025)}, 2025.
\newblock \doi{10.4230/LIPIcs.SEA.2025.16}.
\newblock URL
  \url{https://drops.dagstuhl.de/entities/document/10.4230/LIPIcs.SEA.2025.16}.

\bibitem[Fawzi et~al.(2022)]{alphatensor}
Alhussein Fawzi et~al.
\newblock Discovering faster matrix multiplication algorithms with
  reinforcement learning.
\newblock \emph{Nature}, 610:\penalty0 47--53, 2022.
\newblock \doi{10.1038/s41586-022-05172-4}.
\newblock URL \url{https://doi.org/10.1038/s41586-022-05172-4}.

\bibitem[{Google DeepMind}(2022)]{alphatensorcode}
{Google DeepMind}.
\newblock {AlphaTensor}: Published algorithms and verification artifacts, 2022.
\newblock URL \url{https://github.com/google-deepmind/alphatensor}.
\newblock Accessed September 18, 2026.

\bibitem[{Google DeepMind}(2023)]{gnomecode}
{Google DeepMind}.
\newblock Materials discovery: {GNoME} research artifacts, 2023.
\newblock URL \url{https://github.com/google-deepmind/materials_discovery}.
\newblock Accessed September 18, 2026.

\bibitem[{Google DeepMind}(2024)]{alphageometrycode}
{Google DeepMind}.
\newblock {AlphaGeometry}: Code and model artifacts, 2024.
\newblock URL \url{https://github.com/google-deepmind/alphageometry}.
\newblock Accessed September 18, 2026.

\bibitem[Hadian \& Sobel(1969)Hadian and Sobel]{hadian_sobel}
Abdollah Hadian and Milton Sobel.
\newblock Selecting the {$t$}-th largest using binary errorless comparisons.
\newblock Technical Report Technical Report 121, School of Statistics,
  University of Minnesota, 1969.
\newblock URL \url{https://hdl.handle.net/11299/199105}.

\bibitem[Hubert et~al.(2025)]{alphaproof}
Thomas Hubert et~al.
\newblock Olympiad-level formal mathematical reasoning with reinforcement
  learning.
\newblock \emph{Nature}, 2025.
\newblock \doi{10.1038/s41586-025-09833-y}.
\newblock URL \url{https://www.nature.com/articles/s41586-025-09833-y}.

\bibitem[Kammeyer(2026)]{heisenbergsource}
Holger Kammeyer.
\newblock A counterexample to the determinant approximation conjecture.
\newblock \emph{arXiv preprint arXiv:2609.15567}, 2026.
\newblock URL \url{https://arxiv.org/abs/2609.15567}.

\bibitem[Kassabov et~al.(2026)Kassabov, Landsberg, Souza, and
  Speegle]{centroidsource}
Martin Kassabov, J.~M. Landsberg, Victor Souza, and Philip Speegle.
\newblock Disjoint and nearly disjoint sums of matrix multiplication tensors
  and their centroids.
\newblock \emph{arXiv preprint arXiv:2608.27434}, 2026.
\newblock URL \url{https://arxiv.org/abs/2608.27434}.

\bibitem[Kissinger \& van~de Wetering(2020)Kissinger and van~de Wetering]{pyzx}
Aleks Kissinger and John van~de Wetering.
\newblock Reducing {T}-count with the {ZX}-calculus.
\newblock \emph{Physical Review A}, 102\penalty0 (2):\penalty0 022406, 2020.
\newblock \doi{10.1103/PhysRevA.102.022406}.

\bibitem[Li et~al.(2022)]{alphacode}
Yujia Li et~al.
\newblock Competition-level code generation with {AlphaCode}.
\newblock \emph{Science}, 378:\penalty0 1092--1097, 2022.
\newblock \doi{10.1126/science.abq1158}.
\newblock URL \url{https://arxiv.org/abs/2203.07814}.

\bibitem[Lu et~al.(2024)]{aiscientist1}
Chris Lu et~al.
\newblock The {AI Scientist}: Towards fully automated open-ended scientific
  discovery.
\newblock \emph{arXiv preprint arXiv:2408.06292}, 2024.
\newblock URL \url{https://arxiv.org/abs/2408.06292}.

\bibitem[Merchant et~al.(2023)Merchant, Batzner, Schoenholz, Aykol, Cheon, and
  Cubuk]{gnome}
Amil Merchant, Simon Batzner, Samuel~S. Schoenholz, Muratahan Aykol, Gowoon
  Cheon, and Ekin~Dogus Cubuk.
\newblock Scaling deep learning for materials discovery.
\newblock \emph{Nature}, 624:\penalty0 80--85, 2023.
\newblock \doi{10.1038/s41586-023-06735-9}.
\newblock URL \url{https://www.nature.com/articles/s41586-023-06735-9}.

\bibitem[Novikov et~al.(2025)]{alphaevolve}
Alexander Novikov et~al.
\newblock {AlphaEvolve}: A coding agent for scientific and algorithmic
  discovery.
\newblock \emph{arXiv preprint arXiv:2506.13131}, 2025.
\newblock URL \url{https://arxiv.org/abs/2506.13131}.

\bibitem[{OpenAI}(2026)]{navierstokes2026}
{OpenAI}.
\newblock On the {Navier--Stokes} millennium prize problem, September 2026.
\newblock URL \url{https://openai.com/index/navier-stokes-solution/}.
\newblock Primary research announcement, September 8, 2026, with linked
  mathematical writeup and Lean formalization. Smooth-forcing formulations C
  and D. Accessed September 18, 2026.

\bibitem[{OpenEvolve contributors}(2026)]{openevolve}
{OpenEvolve contributors}.
\newblock {OpenEvolve}: Open-source evolutionary program optimization, 2026.
\newblock URL \url{https://github.com/algorithmicsuperintelligence/openevolve}.
\newblock Pinned experimental revision is reported in the supplementary
  material; accessed September 17, 2026.

\bibitem[{OpenMOSS}(2026)]{agenthpo}
{OpenMOSS}.
\newblock {AgentHPOBench}: A benchmark for evaluating llm agents as sequential
  hyperparameter optimizers, 2026.
\newblock URL \url{https://github.com/OpenMOSS/AgentHPOBench}.
\newblock Official task repository; accessed September 17, 2026.

\bibitem[Romera-Paredes et~al.(2024)]{funsearch}
Bernardino Romera-Paredes et~al.
\newblock Mathematical discoveries from program search with large language
  models.
\newblock \emph{Nature}, 625:\penalty0 468--475, 2024.
\newblock \doi{10.1038/s41586-023-06924-6}.
\newblock URL \url{https://doi.org/10.1038/s41586-023-06924-6}.

\bibitem[Schrittwieser et~al.(2020)Schrittwieser, Antonoglou, Hubert,
  et~al.]{muzero}
Julian Schrittwieser, Ioannis Antonoglou, Thomas Hubert, et~al.
\newblock Mastering {Atari}, {Go}, chess and shogi by planning with a learned
  model.
\newblock \emph{Nature}, 588:\penalty0 604--609, 2020.
\newblock \doi{10.1038/s41586-020-03051-4}.
\newblock URL \url{https://arxiv.org/abs/1911.08265}.

\bibitem[Silver et~al.(2018)Silver, Hubert, Schrittwieser, et~al.]{alphazero}
David Silver, Thomas Hubert, Julian Schrittwieser, et~al.
\newblock A general reinforcement learning algorithm that masters chess, shogi,
  and {Go} through self-play.
\newblock \emph{Science}, 362\penalty0 (6419):\penalty0 1140--1144, 2018.
\newblock \doi{10.1126/science.aar6404}.
\newblock URL \url{https://doi.org/10.1126/science.aar6404}.

\bibitem[Sinclair(2026)]{hbetasource}
Thomas Sinclair.
\newblock The radial derivative on the graded {M\"obius} algebra.
\newblock \emph{arXiv preprint arXiv:2608.18519}, 2026.
\newblock URL \url{https://arxiv.org/abs/2608.18519}.

\bibitem[Trinh et~al.(2024)Trinh, Wu, Le, He, and Luong]{alphageometry}
Trieu~H. Trinh, Yuhuai Wu, Quoc~V. Le, He~He, and Thang Luong.
\newblock Solving olympiad geometry without human demonstrations.
\newblock \emph{Nature}, 625:\penalty0 476--482, 2024.
\newblock \doi{10.1038/s41586-023-06747-5}.
\newblock URL \url{https://www.nature.com/articles/s41586-023-06747-5}.

\bibitem[Yamada et~al.(2025)]{aiscientist2}
Yutaro Yamada et~al.
\newblock The {AI Scientist-v2}: Workshop-level automated scientific discovery
  via agentic tree search.
\newblock \emph{arXiv preprint arXiv:2504.08066}, 2025.
\newblock URL \url{https://arxiv.org/abs/2504.08066}.

\end{thebibliography}
\bibliographystyle{iclr2027_conference}
\expandafter\enddocument
\fi

\end{document}